\documentclass[11pt]{article}

\usepackage[utf8]{inputenc}
\usepackage[T1]{fontenc}
\usepackage{times}
\usepackage{microtype}

\usepackage[top=1in, bottom=1in, left=1.25in, right=1.25in]{geometry}

\usepackage{amsmath, amssymb}

\usepackage{graphicx}
\graphicspath{{figures/}}

\usepackage{booktabs}
\usepackage{array}
\usepackage{tabularx}
\usepackage{multirow}
\usepackage{longtable}

\usepackage[font=small, labelfont=bf, format=plain, justification=justified]{caption}

\usepackage{xcolor}

\usepackage[colorlinks=true, citecolor=blue, linkcolor=blue, urlcolor=blue,
            pdftitle={Knowledge-augmented Agentic AI for Mental Health Medication Information Seeking},
            pdfauthor={Author et al.}]{hyperref}

\usepackage{setspace}
\usepackage{indentfirst}
\usepackage{titlesec}
\titleformat{\section}{\large\bfseries}{}{0pt}{}
\titleformat{\subsection}{\normalsize\bfseries\itshape}{}{0pt}{}
\titleformat{\subsubsection}[runin]{\normalsize\bfseries}{}{0pt}{}[\quad]
\titleformat{\paragraph}{\normalsize\bfseries}{}{0pt}{}
\titlespacing*{\section}{0pt}{14pt}{4pt}
\titlespacing*{\subsection}{0pt}{10pt}{3pt}
\titlespacing*{\subsubsection}{0pt}{6pt}{0pt}
\titlespacing*{\paragraph}{0pt}{6pt}{2pt}

\usepackage{abstract}

\usepackage{placeins}
\usepackage{float}

\usepackage{fancyvrb}
\usepackage{tcolorbox}
\tcbuselibrary{breakable, skins}

\newcommand{\appendixcontentsline}[3]{%
  \noindent\hyperref[#1]{\textbf{#2}\quad #3}\dotfill\pageref{#1}\par}
\newcommand{\appendixcontentsitem}[3]{%
  \noindent\hspace*{1.5em}\hyperref[#1]{#2\quad #3}\dotfill\pageref{#1}\par}

\usepackage[numbers, compress, sort]{natbib}

\usepackage{ifthen}

\newcommand{\affmark}[1]{\textsuperscript{#1}}
\newcommand{\corrmark}{\textsuperscript{*}}
\newcommand{\equalmark}{\textsuperscript{\ensuremath{\dagger}}}
\begin{document}
\thispagestyle{empty}

% ── Title ───────────────────────────────────────────────────────────────────
\begin{flushleft}
\sffamily
{\fontsize{16.5}{20}\selectfont\bfseries
Knowledge-augmented Agentic AI for Mental Health\\
Medication Information Seeking
\par}

\vspace{10pt}

% ── Authors ─────────────────────────────────────────────────────────────────
{\normalsize\bfseries
Huizi Yu\affmark{1}\equalmark,
Jian Liu\affmark{2}\equalmark,
Wenkong Wang\affmark{3}\equalmark,
Lingyao Li\affmark{4},
Jiayan Zhou\affmark{5},
Zhaoqian Xue\affmark{6},\\
Xiang Li\affmark{2},
Xinxin Lin\affmark{2},
Zhiying Liang\affmark{2},
Zhuoru Wu\affmark{2},
Siyuan Ma\affmark{7},
Xin Ma\affmark{3}, and
Lizhou Fan\affmark{2,8}\corrmark
\par}

\vspace{6pt}

% ── Affiliations ─────────────────────────────────────────────────────────────
{\small
\affmark{1}Department of Medicine and Therapeutics, The Chinese University of Hong Kong, Shatin, Hong Kong SAR, China\\
\affmark{2}Department of Psychiatry, The Chinese University of Hong Kong, Shatin, Hong Kong SAR, China\\
\affmark{3}School of Control Science and Engineering, Shandong University, Ji'nan, Shandong, China\\
\affmark{4}College of Information Science, University of Arizona, Tucson, Arizona, USA\\
\affmark{5}Department of Medicine, Stanford University School of Medicine, Stanford University, Palo Alto, California, USA\\
\affmark{6}Perelman School of Medicine, The University of Pennsylvania, Philadelphia, Pennsylvania, USA\\
\affmark{7}Department of Biostatistics, Vanderbilt University, Nashville, Tennessee, USA\\
\affmark{8}Li Ka Shing Institute of Health Sciences, Faculty of Medicine, The Chinese University of Hong Kong, Shatin, N.T., Hong Kong SAR, China\\

\vspace{12pt}

\equalmark These authors contributed equally.\\
\corrmark Correspondence to: Lizhou Fan (\href{mailto:leofan@cuhk.edu.hk}{leofan@cuhk.edu.hk})
\par}
\end{flushleft}

\vspace{12pt}

% ────────────────────────────────────────────────────────────────────────────
\noindent{\large\sffamily\bfseries ABSTRACT\par}
\vspace{8pt}
\noindent{\sffamily
Patients increasingly seek medication information online, yet safety knowledge for psychiatric
drugs is split between regulatory adverse-event records, which are authoritative but abstract,
and patient narratives, which are experience-near but unvalidated. Integrating them without
conflating evidence and anecdote is especially consequential in psychiatry, where poorly
contextualised information can amplify fear, nocebo responses, and non-adherence. Here we
develop a provenance-aware, knowledge-graph-based multi-agent framework unifying 466,525 Reddit
posts, 60,782 WebMD reviews, and twenty years of U.S. FDA Adverse Event Reporting System records
for nine antidepressants. A large-language-model entity-recognition pipeline benchmarked against
physician annotations reached highest F1 scores of 0.969 for medications and 0.973 for conditions.
The two community platforms were far more concordant with each other (overlap up to a Jaccard
similarity of 0.905) than with regulatory reports, indicating that patient-generated data form a
partly independent safety signal. For sertraline, many adverse events appeared in community
sources hundreds of days before the corresponding FDA date. A Neo4j knowledge graph grounded in
ATC-N, ICD-10, and MedDRA vocabularies preserves provenance, keeping every claim traceable and
regulatory facts distinct from patient experience. These results establish source-aware
integration as a route to more auditable psychiatric medication information, with usefulness and
patient benefit to be tested prospectively.

\par}

\setcounter{page}{1}

% ════════════════════════════════════════════════════════════════════════════
\section*{Introduction}
% ════════════════════════════════════════════════════════════════════════════

Psychiatric pharmacotherapy is often long term, individualized, and dynamically adjusted over the
course of illness, which makes patient understanding of medication benefits, adverse effects, and
uncertainty especially important. Yet in routine care, medication counseling is frequently
compressed into short clinical encounters, leaving many patients with unanswered questions after
the visit. This gap matters in a digital environment where seeking health information online has
become routine across regions: in Europe, 55\% of people aged 16--74 sought health-related
information online in 2020~\cite{ref1}; in the United States, the proportion of adults who used
the internet first for their most recent health information search increased from 61.2\% in 2008 to
74.4\% in 2017~\cite{ref2}; and in Asia, a 10-country survey found that 71.6\% of smartphone users
sought health information on their smartphones at least a few times per month~\cite{ref3}. Prior
syntheses suggest that online health information seeking is closely tied to medication-related
decision-making, even if its relationship with adherence is complex and heterogeneous across
settings~\cite{ref4,ref5}. In parallel, recent work shows that patients' knowledge of newly prescribed
medications remains incomplete, particularly for practical details such as administration and side
effects, underscoring the need for clearer and more accessible medication education
tools~\cite{ref6}.

The challenge may be particularly acute in psychiatric care. Patients prescribed antidepressants,
antipsychotics, mood stabilizers, or anxiolytics often seek explanations not only about formal
adverse-effect lists, but also about how medication-related experiences unfold in daily life:
changes in sleep, appetite, concentration, affect, weight, motivation, or discontinuation
symptoms. Peer forums and online communities can provide validation, practical language, and a
sense of not being alone; recent mixed-methods work in mental health forums suggests that users
often derive emotional support, normalization, and practical benefit from these spaces~\cite{ref7}.
At the same time, peer narratives are highly heterogeneous in accuracy and representativeness. In
medication contexts, expectations can themselves shape the experience of side effects, and negative
or poorly contextualized information may amplify anxiety and nocebo
responses~\cite{ref8,ref9}. For psychiatric medications, where adherence is already vulnerable to
fear, stigma, and uncertainty, this creates a communication problem: patients need information that
is understandable and experience-near, but they also need it to be proportionate, evidence-aware,
and safe.

Existing information channels each address only part of this need. Regulatory and professional
sources offer standardized, evidence-based descriptions of indications, warnings, and reported
adverse events, but they are often experienced by patients as abstract, decontextualized, or
difficult to map onto lived experience. FAERS, for example, is a major postmarketing safety
resource explicitly designed to support the FDA's postmarketing surveillance of marketed drugs and
biologics~\cite{ref10}. Meanwhile, social media and consumer-generated narratives can surface
patient-reported experiences that may be absent, delayed, or underemphasized in formal channels.
A recent scoping review concluded that social media analysis may serve as a useful supplementary
source for adverse-event detection and pharmacovigilance, while also emphasizing the need for
careful validation and source-aware interpretation~\cite{ref11}. The central problem, therefore,
is not whether authoritative data or patient narratives are ``better,'' but how to integrate them
so that official evidence remains the factual backbone while real-world experience adds context
rather than distortion.

Large language models offer a potentially powerful interface for this integration because they can
translate complex medical information into fluent, user-facing explanations and support
conversational querying. However, the same properties that make LLMs accessible also make them
risky in health settings. Systematic reviews show rapid growth in LLM applications for patient
care and chatbot-based health advice, but also document recurring concerns around factual
reliability, transparency, evaluation quality, and governance~~\cite{ref12,ref13,ref19}. More pointedly,
benchmark work in clinical decision-making shows that current LLMs are not ready for autonomous
clinical use~\cite{ref14}, and workflow-level evaluations further suggest that clinical readiness
depends on how models gather, interpret, and communicate information across multi-turn consultation
processes, not only final-answer accuracy~\cite{ref20}. Recent safety-focused research has called for clinician-centered frameworks to assess hallucinations and clinical harm rather than relying on surface plausibility alone~\cite{ref15}, while evidence-guided alignment frameworks have also
been proposed to improve structured psychiatric clinical reasoning in lighter-weight LLMs~\cite{ref22}.These concerns are especially salient in psychiatry, where poorly framed
answers about adverse effects, suicidality, withdrawal, or vulnerable populations could increase
fear, disrupt adherence, or undermine trust. As a result, any AI system for psychiatric medication
education should be explicitly constrained, source-grounded, and evaluated as an educational aid
rather than a substitute for professional judgment~\cite{ref16}.

A promising way to achieve this is to combine LLMs with knowledge graphs and multi-agent
orchestration. Knowledge graphs provide a structured representation of entities, relations, and
provenance across heterogeneous information sources, improving interpretability and supporting
evidence-aware retrieval~\cite{ref17}. In healthcare, retrieval-augmented and graph-grounded LLM
workflows are increasingly used to reduce hallucination and improve factual consistency by
separating evidence retrieval from answer generation~\cite{ref15}. For the present application,
this architecture is particularly attractive because it can encode different evidence roles within
the same system: regulatory or professional sources can anchor factual claims, while community
narratives can be surfaced as contextualized, secondary experience signals. A multi-agent design can then distribute functions such as query understanding, source selection,
retrieval, synthesis, and validation, thereby reinforcing internal checks and clearer safety
boundaries~\cite{ref15,ref16,ref18,ref21,ref23,ref24}.

Against this background, the present study constructs and characterises a knowledge-graph-driven
multi-agent AI system for psychiatric drug adverse event information using multi-source health
data. The system integrates community and regulatory information from Reddit, WebMD, and
FDA/FAERS into a unified knowledge framework designed to support educational question answering
about medication use, side effects, withdrawal, and risk communication, rather than diagnosis or
treatment recommendation. The core contributions of this work are: (i)~a scalable LLM-based NER
pipeline benchmarked across nine state-of-the-art models on physician-annotated data; (ii)~a
multi-source comparative analysis of adverse-event profiles across regulatory and
community-derived data streams, including cross-source overlap, frequency structure, and
temporal lead-time analysis for nine antidepressants; and (iii)~a Neo4j knowledge graph and
multi-agent chatbot architecture providing provenance-aware, safety-constrained retrieval of
psychiatric medication information.

% ════════════════════════════════════════════════════════════════════════════
\section*{Methods}
% ════════════════════════════════════════════════════════════════════════════

\subsection*{Drug List Construction (ATC-N)}

We built the drug dictionary from the WHO Anatomical Therapeutic Chemical (ATC) system, which
hierarchically classifies medicines by anatomical system, therapeutic/pharmacologic class, and
generic names (i.e., chemical substance). From the ATC Index, we programmatically retrieved the
hierarchy and filtered to the N (Nervous system) branch, which contained 626 generic drug names.
To improve recall during downstream text mining, we expanded generic names to brand synonymy using
a constrained LLM prompt that returned brand names in a simple structured format (see
Appendix~\ref{app:A}). The resulting ATC-N dictionary---containing ATC code, generic drug name,
and curated brand names---serves as the basis for keyword generation in data collection and for
medication normalization across community and agency sources.

\subsection*{Data Collection (Community and Agency)}

We collected community-based discourse posts from Reddit using the Academic Torrents public
dataset, restricting the analysis window to June 2005 through April 2025. Reddit is a large,
topic-organized social platform where users share timestamped narratives within public forums. We
included submissions whose title or body contained any ATC-N generic or curated brand keyword. To
enhance textual quality and specificity, we removed entries with missing content, concatenated
titles and body text, excluded records with fewer than 10 words, and deduplicated exact
(title,~body) pairs. Posts triggered solely by ambiguous keyword matches (e.g., ``lithium'',
``cocaine'') were also excluded. Finally, we applied automated language detection to retain only
English-language posts. This process resulted in 1,138,331 unique, English, keyword-positive
Reddit posts. Upon closer inspection, we found that a substantial portion of Reddit posts retained
after keyword- and rule-based filtering still lacked meaningful clinical information. To further
enhance the overall data quality, we introduced a secondary filtering stage designed to
differentiate information-rich from information-poor content (see Appendix~\ref{app:B}). The
final Reddit post collection contains 466,525 posts.

For the second community-based discourse source, we collected structured medication reviews from
WebMD, a consumer health platform where patients and caregivers submit rated, time-stamped
narratives about prescribed drugs. For each ATC-N generic name, we identified the corresponding
review page(s) and scraped all available entries, capturing users' age group, gender, duration of
use, reviewer role (patient or caregiver), overall rating, and free-text review. Compared with
Reddit, WebMD provides semi-structured demographic and usage fields alongside narrative text,
enabling complementary community-based signal characterization and stratified analyses. We
completed the same rule-based and model-based filtering as the Reddit data, and the remaining
WebMD corpus contained 60,782 reviews.

We analyzed the U.S. FDA's Adverse Event Reporting System (FAERS), a large, long-standing
national pharmacovigilance database with millions of post-marketing reports and broad regulatory
prestige. We aligned the analytic window from 2005 Quarter~2 to 2025 Quarter~1 to correspond with
community data. Using the open-source \textit{faers-toolkit} (GitHub), we parsed quarterly
releases into SQLite. We then filtered FAERS to the ATC-N drug set by matching each report's drug
names against our ATC-N dictionary. Finally, we restricted records by FDA receipt date and applied
light normalization (trimming whitespace, converting empty fields to null). All reporter types
(manufacturer, clinician, consumer) were retained to maximize coverage.

\subsection*{LLM-based Named Entity Recognition (NER)}
\label{sec:ner}

To extract structured clinical information from unstructured Reddit and WebMD posts, we employed
a large language model (LLM)-based NER pipeline. Each post was processed using a single-pass
structured extraction prompt (see Appendix~\ref{app:D}) instructing the model to identify and
return a JSON object containing: (i)~medications with dosage, dosage form, duration of use, and
continuation status; (ii)~primary psychiatric condition with severity and diagnostic status;
(iii)~comorbid conditions; and (iv)~side effects with associated drug, severity, frequency, and
duration. Relations between entities (TREATS, CAUSES, CAUSES\_BY\_WITHDRAW, COMORBID\_WITH)
were also extracted in the same pass.

We benchmarked nine state-of-the-art LLMs on a physician-annotated gold-standard NER dataset
(annotation interface shown in Appendix~\ref{app:C}; model comparison reported in the Results).
GPT-4.1-mini was selected as the pipeline default based on its favourable balance of extraction
accuracy, throughput, and cost (Appendix~\ref{app:F}).

\subsection*{Entity Mapping and Knowledge Graph Construction}
\label{sec:kg-construction}

\paragraph{Entity canonicalization.}
Raw entity strings extracted by the NER pipeline were mapped to controlled biomedical vocabularies
to ensure consistency across posts. Medications were aligned to ATC-N ingredient-level identifiers
using our ATC-N dictionary. Conditions (primary and comorbid) were mapped to International
Classification of Diseases 10th Revision (ICD-10) terms, and side effects were mapped to Medical
Dictionary for Regulatory Activities (MedDRA) Preferred Terms. Mapping was performed by
embedding-based nearest-neighbour retrieval using \textit{text-embedding-3-small}, with
entity-type-specific cosine-similarity thresholds calibrated to maximize Youden's J statistic on
a physician-annotated gold standard (see Appendix~\ref{app:E}).

\paragraph{Graph schema.}
The knowledge graph was implemented in Neo4j with four main node types: Post, Medication,
Condition, and SideEffect. Post nodes are lightweight
anchor nodes containing only a unique identifier and minimal metadata; full post text is stored in
a linked SQLite sidecar database with full-text search indexing (see Appendix~\ref{app:G}).
Domain entity nodes store canonicalized ontology-level identifiers and are deduplicated via
Neo4j uniqueness constraints, so all lexical variants of the same clinical concept collapse into a
single node.

\paragraph{Typed edges.}
Relations were materialized as four typed edge classes: \textsc{TREATS} (medication--condition),
\textsc{CAUSES} (medication--side effect), \textsc{CAUSES\_BY\_WITHDRAW} (medication--side
effect, discontinuation context), and \textsc{COMORBID\_WITH} (condition--condition). In
addition, \textsc{MENTIONS} edges link each Post node to all domain entities referenced in that
post, preserving direct provenance linkage for evidence-traced retrieval. Where the same entity
pair appeared across multiple posts, supporting post identifiers were accumulated as an edge
property list.

\subsection*{Multi-source Comparative Analysis}

To compare AE patterns across data sources, we conducted a multi-source analysis integrating FDA,
WebMD, and Reddit data for a set of antidepressant drugs. After source-specific preprocessing and
AE extraction, records from the three sources were aligned at the drug--AE level using harmonized
AE terms so that comparable adverse effects could be evaluated across platforms. This allowed us
to examine both shared and source-specific AE patterns across regulatory and community-derived
data.

We assessed cross-source similarity from several complementary perspectives. First, we measured
the overlap in AE profiles across source pairs to evaluate the extent to which the same adverse
effects were represented in different datasets. Second, we summarized how evenly AE information
was distributed across the three sources, allowing us to distinguish drugs with relatively
balanced cross-source representation from those whose AE profiles were concentrated more strongly
in one or two sources. Third, we compared the relative prominence of AEs across sources to assess
whether some adverse effects were emphasized more strongly in one source than another.

We also examined temporal differences across sources using dated AE records from FDA, WebMD, and
Reddit. After aligning AE terms across the three datasets, we identified when each adverse effect
first appeared in each source and compared the timing of emergence across regulatory and
community-derived data. This temporal analysis was used to assess whether some adverse effects
tended to appear earlier in community reporting streams than in FDA, or vice versa.

Overall, this comparative framework allowed us to evaluate similarity and divergence across FDA,
WebMD, and Reddit at multiple levels, including AE overlap, relative prominence, source balance,
and timing of first appearance. The goal was not to treat these data streams as interchangeable,
but to characterize how they converge and differ in representing antidepressant adverse effects.

\subsection*{Multi-Agent Adverse Event Chatbot}
\label{sec:chatbot-method}

\begin{figure}[htbp]
  \centering
  \includegraphics[width=0.95\linewidth]{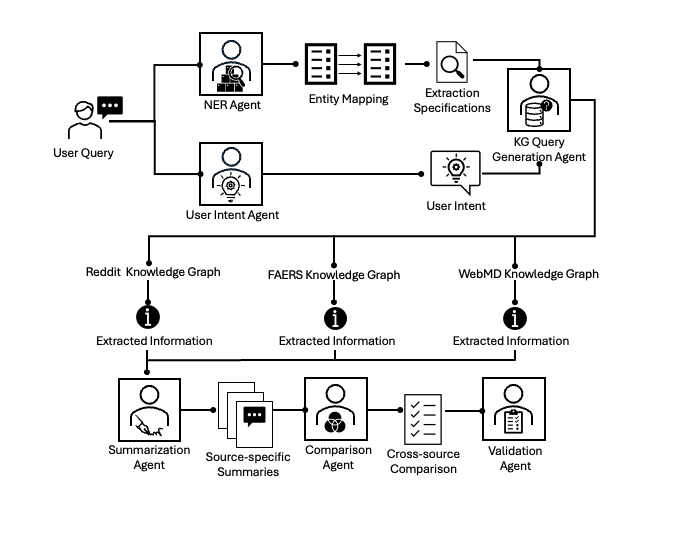}
  \caption{Multi-agent adverse-event information-seeking architecture. The system receives a user
  query and routes it through two parallel analysis streams: (i)~a \textit{NER Agent} that
  extracts medication entities and maps them to canonical identifiers via Entity Mapping, producing
  Extraction Specifications; and (ii)~a \textit{User Intent Agent} that parses the clinical
  question type. Both outputs converge at the \textit{KG Query Generation Agent}, which formulates
  structured queries against three parallel knowledge graphs (Reddit KG, FAERS KG, and WebMD KG).
  Extracted information from each source is then passed to the \textit{Summarization Agent}, which
  produces source-specific summaries. The \textit{Comparison Agent} synthesizes these into a
  cross-source comparison, which is reviewed by the \textit{Validation Agent} before a final
  evidence-grounded response is returned to the user.}
  \label{fig:multiagent}
\end{figure}

The chatbot follows a multi-agent pipeline designed to decompose information seeking into
specialist sub-tasks while enforcing source attribution throughout (Fig.~\ref{fig:multiagent}),
consistent with recent role-structured agentic workflows in healthcare, mental health, emergency
medical services, and simulated-patient systems~\cite{ref21,ref23,ref24}.
When a user submits a query,
two agents process it in parallel: a \textit{NER Agent} identifies all medication names and maps
them to canonical ATC-N identifiers via the entity-mapping module, producing a structured
extraction specification; and a \textit{User Intent Agent} classifies the clinical question type
(e.g., general adverse event inquiry, demographic-stratified question, or longitudinal trend
query). The extraction specification and the parsed user intent are combined by the \textit{KG
Query Generation Agent}, which selects the appropriate knowledge graphs and formulates
Cypher queries against each source (Reddit KG, FAERS KG, WebMD KG). Source selection is
intent-driven: demographic or epidemiological questions route primarily to the FAERS and WebMD
graphs, which carry structured age, sex, and temporal metadata; experiential or context-rich
questions route preferentially to the Reddit graph.

Retrieved evidence from each graph is passed to the \textit{Summarization Agent}, which
produces independent source-specific summaries preserving provenance tags. The
\textit{Comparison Agent} then synthesizes these summaries into a cross-source comparison
that highlights consensus and divergence across community and regulatory data. Finally, the
\textit{Validation Agent} reviews the synthesized response against a predefined safety ruleset
before the answer is returned to the user. The architecture is designed to constrain generation
to retrieved graph context, thereby reducing hallucination risk while preserving user-facing
fluency---a property particularly important for psychiatric medication information, where
unsupported claims about adverse effects, discontinuation, or drug interactions could amplify
nocebo responses or disrupt adherence.

% ════════════════════════════════════════════════════════════════════════════
\section*{Results}
% ════════════════════════════════════════════════════════════════════════════

\subsection*{Evaluation of NER Performance}
\label{sec:ner-results}

Across all nine LLMs evaluated on the physician-annotated NER benchmark, performance varied
substantially by entity category and attribute type. For medication-related entities
(Table~\ref{tab:table1}), GPT-4.1-mini achieved the highest F1 for medication name extraction
(0.969), followed by Claude-Sonnet-4 (0.952) and Gemini-2.5-Flash (0.947). Dosage attribute
extraction was markedly harder across all models, with scores ranging from 0.523 (GPT-5-nano) to
0.751 (Claude-Sonnet-4), likely reflecting the heterogeneous ways in which patients report drug
doses in unstructured narratives. Dosage form extraction was near-ceiling for most premium-tier
models ($>$0.98), consistent with its relatively stereotyped surface forms in patient text.

For condition-related entities (Table~\ref{tab:table2}), Claude-Sonnet-4 achieved the highest
primary-condition F1 (0.973), with GPT-4.1-mini and GPT-4o-mini both reaching 0.966. Comorbid
condition extraction was uniformly lower than primary-condition extraction across all models,
reflecting the greater syntactic ambiguity and higher clinical density of posts describing multiple
co-occurring diagnoses. Duration of illness showed the widest cross-model variability
(range 0.222--0.562), indicating that temporal expressions in patient narratives remain a
persistent challenge for current LLMs.

For side-effect entities (Table~\ref{tab:table3}), Deepseek-V3 achieved the highest
side-effect name F1 (0.912), followed by Claude-Sonnet-4 (0.879) and Gemini-2.5-Flash (0.846).
Attribute-level extraction for side effects---particularly duration (range 0.188--0.476) and
frequency (range 0.125--0.750)---was substantially lower than name-level F1, highlighting the
tendency of patients to describe adverse events qualitatively rather than with explicit temporal
or recurrence characterization.

Balancing NER accuracy across all three entity categories against deployment throughput and cost
(see Appendix~\ref{app:F}), GPT-4.1-mini was selected as the pipeline default for full-corpus
extraction. The consistent gap between name-level and attribute-level extraction across entity
types underscores a fundamental challenge in clinical NLP from patient-generated text: while
models reliably detect that an adverse event occurred, precise characterization of its severity,
timing, and recurrence requires either richer contextual framing or supplementary annotation.

\begin{table}[htbp]
\centering
\caption{Named-entity recognition performance of LLMs on medication-related entity extraction (F1 scores).}
\label{tab:table1}
\small
\begin{tabular}{lccc}
\toprule
\textbf{Model} & \textbf{Medication} & \textbf{Dosage} & \textbf{Dosage Form} \\
\midrule
GPT-5-mini       & 0.936 & 0.671 & 0.994 \\
GPT-5-nano       & 0.893 & 0.523 & 0.985 \\
GPT-4.1-mini     & \textbf{0.969} & 0.746 & \textbf{0.997} \\
GPT-4.1-nano     & 0.915 & 0.641 & 0.848 \\
GPT-4o-mini      & 0.902 & 0.717 & 0.978 \\
Claude-Sonnet-4  & 0.952 & \textbf{0.751} & 0.991 \\
Gemini-2.5-Flash & 0.947 & 0.738 & 0.984 \\
Deepseek-V3      & 0.961 & 0.691 & 0.953 \\
Qwen3-235b-A22b  & 0.936 & 0.693 & 0.962 \\
\bottomrule
\end{tabular}
\end{table}

\begin{table}[htbp]
\centering
\caption{Named-entity recognition performance of LLMs on condition-related entity extraction (F1 scores).}
\label{tab:table2}
\small
\begin{tabular}{lccccc}
\toprule
\textbf{Model} & \textbf{Condition} & \textbf{Comorbid} & \textbf{Diagnosed} & \textbf{Severity} & \textbf{Duration} \\
\midrule
GPT-5-mini       & 0.963 & 0.862 & 0.837 & 0.714 & 0.357 \\
GPT-5-nano       & 0.959 & 0.839 & 0.774 & 0.375 & 0.222 \\
GPT-4.1-mini     & 0.966 & \textbf{0.926} & 0.839 & \textbf{0.833} & 0.500 \\
GPT-4.1-nano     & 0.929 & 0.827 & 0.692 & 0.545 & 0.500 \\
GPT-4o-mini      & 0.966 & 0.875 & 0.860 & 0.714 & 0.333 \\
Claude-Sonnet-4  & \textbf{0.973} & 0.925 & \textbf{0.872} & 0.750 & \textbf{0.562} \\
Gemini-2.5-Flash & 0.913 & 0.836 & 0.840 & 0.818 & 0.429 \\
Deepseek-V3      & 0.965 & 0.857 & 0.773 & 0.818 & 0.533 \\
Qwen3-235b-A22b  & 0.920 & 0.882 & 0.864 & 0.667 & 0.438 \\
\bottomrule
\end{tabular}
\end{table}

\begin{table}[htbp]
\centering
\caption{Named-entity recognition performance of LLMs on side-effect entity extraction (F1 scores).}
\label{tab:table3}
\small
\begin{tabular}{lcccc}
\toprule
\textbf{Model} & \textbf{Side Effect} & \textbf{Severity} & \textbf{Duration} & \textbf{Frequency} \\
\midrule
GPT-5-mini       & 0.837 & 0.690 & \textbf{0.476} & 0.667 \\
GPT-5-nano       & 0.715 & 0.417 & 0.462 & 0.200 \\
GPT-4.1-mini     & 0.858 & 0.571 & 0.389 & 0.714 \\
GPT-4.1-nano     & 0.765 & 0.636 & 0.300 & 0.667 \\
GPT-4o-mini      & 0.749 & 0.395 & 0.188 & 0.125 \\
Claude-Sonnet-4  & 0.879 & 0.732 & 0.350 & 0.500 \\
Gemini-2.5-Flash & 0.846 & \textbf{0.742} & 0.333 & \textbf{0.750} \\
Deepseek-V3      & \textbf{0.912} & 0.641 & 0.429 & 0.333 \\
Qwen3-235b-A22b  & 0.800 & 0.514 & 0.353 & 0.143 \\
\bottomrule
\end{tabular}
\end{table}

\FloatBarrier

\subsection*{Knowledge Graph}
\label{sec:kg-results}

\begin{figure}[htbp]
  \centering
  \includegraphics[width=0.95\linewidth]{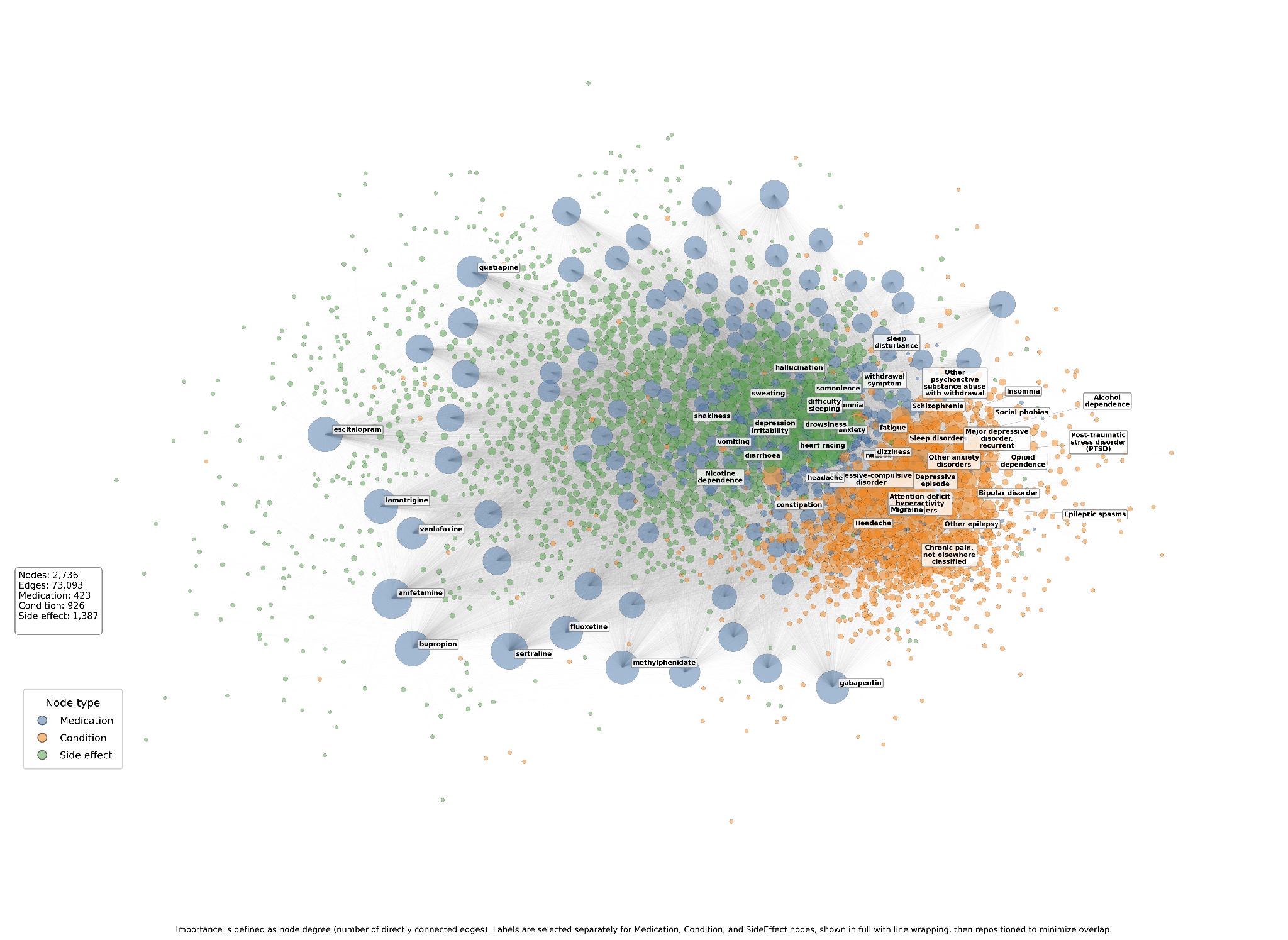}
  \caption{Overview of the Reddit-derived mental health knowledge graph. Nodes represent
  canonicalized clinical entities (\texttt{Medication}, \texttt{Condition}, \texttt{SideEffect};
  color-coded by type) or source \texttt{Post}s. Typed edges encode treatment relationships
  (\textsc{TREATS}), adverse-effect associations (\textsc{CAUSES}, \textsc{CAUSES\_BY\_WITHDRAW}),
  comorbid-disease co-occurrence (\textsc{COMORBID\_WITH}), and source-post provenance
  (\textsc{MENTIONS}). The layout highlights hub nodes corresponding to commonly prescribed
  antidepressants and frequently reported psychiatric diagnoses.}
  \label{fig:kg-overview}
\end{figure}

The Reddit-derived mental health knowledge graph integrates ontology-grounded representations of
psychiatric pharmacotherapy from 466,525 information-rich posts (Fig.~\ref{fig:kg-overview}).
Medications were canonicalized
to ATC-N ingredient-level identifiers, conditions to ICD-10 terms, and side effects to MedDRA
Preferred Terms using embedding-based nearest-neighbour matching with entity-type-specific
cosine-similarity thresholds calibrated to maximize Youden's J statistic on a physician-annotated
gold standard (see Appendix~\ref{app:E}). The graph encodes four main typed edge
classes---\textsc{TREATS}, \textsc{CAUSES}, \textsc{CAUSES\_BY\_WITHDRAW}, and
\textsc{COMORBID\_WITH}---along with \textsc{MENTIONS} edges that preserve direct provenance
linkage between entity nodes and the Reddit posts in which they appear, enabling evidence-traced
retrieval for downstream question answering.

\begin{figure}[htbp]
  \centering
  \includegraphics[width=0.84\linewidth]{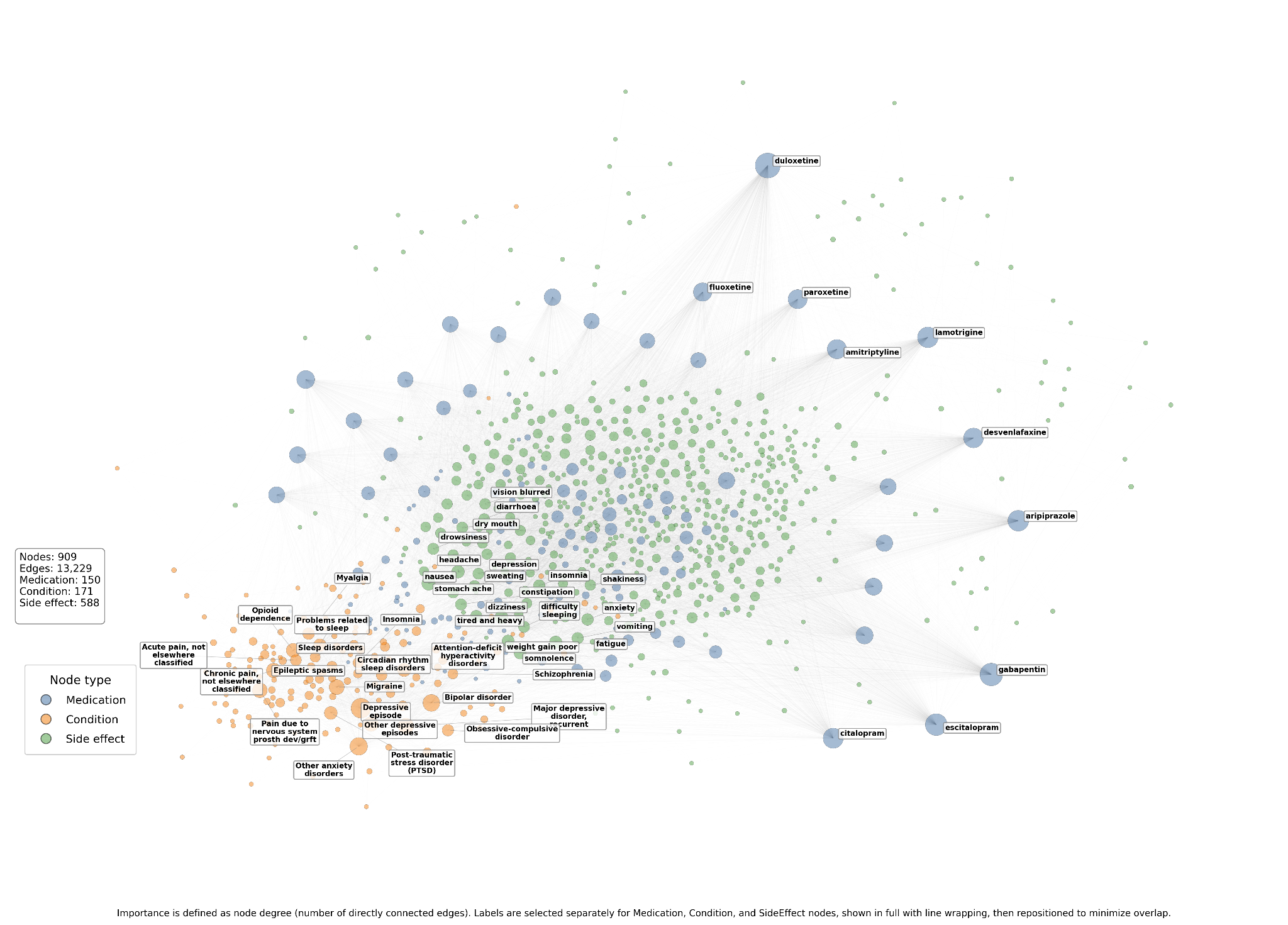}
  \caption{Sertraline-centred subgraph linking the medication to co-occurring ICD-10 conditions,
  MedDRA adverse effects and supporting Reddit posts. Edge thickness is proportional to the
  number of supporting posts; \textsc{TREATS}, \textsc{CAUSES} and \textsc{COMORBID\_WITH}
  denote treatment, adverse-event and comorbidity relations.}
  \label{fig:kg-detail}
\end{figure}

The design deliberately separates entity-level semantics from post-level content: full post text
is stored in a linked SQLite sidecar database with full-text search indexing, keeping the core
knowledge graph compact and privacy-preserving while preserving the ability to trace any inferred
relation back to its source posts for downstream verification and retrieval-augmented generation.
A sertraline-centred subgraph illustrates how medication nodes connect to co-occurring
conditions, adverse events, and supporting Reddit posts (Fig.~\ref{fig:kg-detail}).

\subsection*{Similarity Across Sources}
\label{sec:similarity}

\begin{figure}[htbp]
  \centering
  \includegraphics[width=0.75\linewidth]{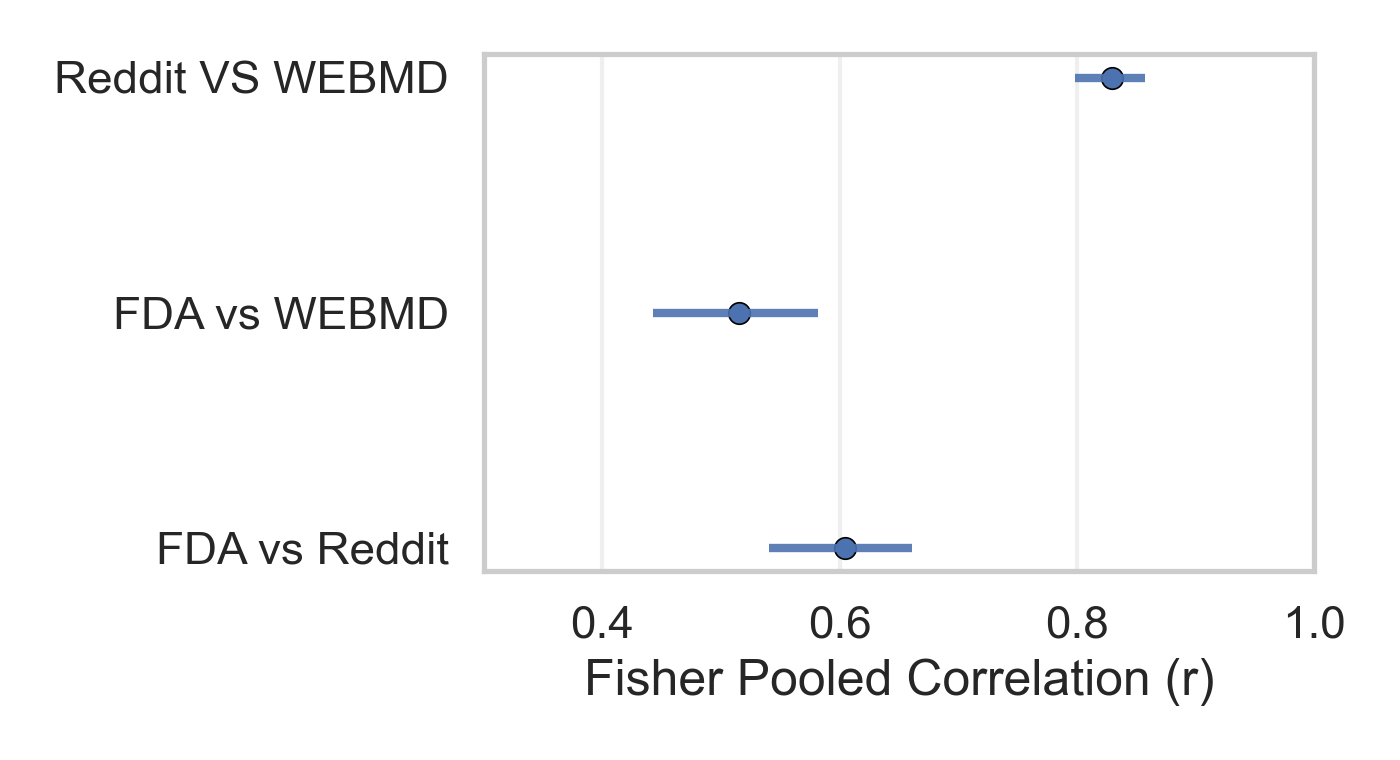}
  \caption{Pooled adverse-event profile correlations across FDA, WebMD, and Reddit for
  antidepressants. Points show Fisher-pooled correlations across the nine-drug panel; horizontal
  bars indicate uncertainty intervals from the pooled estimates.}
  \label{fig:ae-overview}
\end{figure}

We first compared adverse-event (AE) similarity across FDA, WebMD, and Reddit for nine
antidepressants using pairwise Jaccard indices and source-balance metrics
(Fig.~\ref{fig:ae-overview}; Table~\ref{tab:table4}). Across drugs, WebMD
and Reddit generally showed the greatest overlap, with WebMD--Reddit Jaccard values reaching
0.905 for desvenlafaxine, 0.875 for duloxetine, 0.833 for fluoxetine, 0.805 for paroxetine, and
0.786 for sertraline. In contrast, overlap between FDA and the two community-derived sources was
typically lower, although the extent of this difference varied by drug. Desvenlafaxine showed
high concordance across all three source pairs, whereas vilazodone showed marked asymmetry, with
high FDA--Reddit overlap (0.756) but much lower FDA--WebMD and WebMD--Reddit overlap (0.286 and
0.293, respectively). Together, these findings suggest that a shared AE core exists across
sources, but the degree of cross-source agreement is drug-specific. Source-balance metrics showed
that overlap in AE identity and balance in AE volume were not identical properties. Fluoxetine,
phenelzine, sertraline, and desvenlafaxine had relatively high composition entropy and evenness,
indicating a more balanced distribution of AE counts across sources. By contrast, duloxetine and
amitriptyline showed lower entropy and evenness, suggesting that their AE profiles were more
concentrated in one or two sources.

\begin{figure}[htbp]
  \centering
  \includegraphics[width=0.95\linewidth]{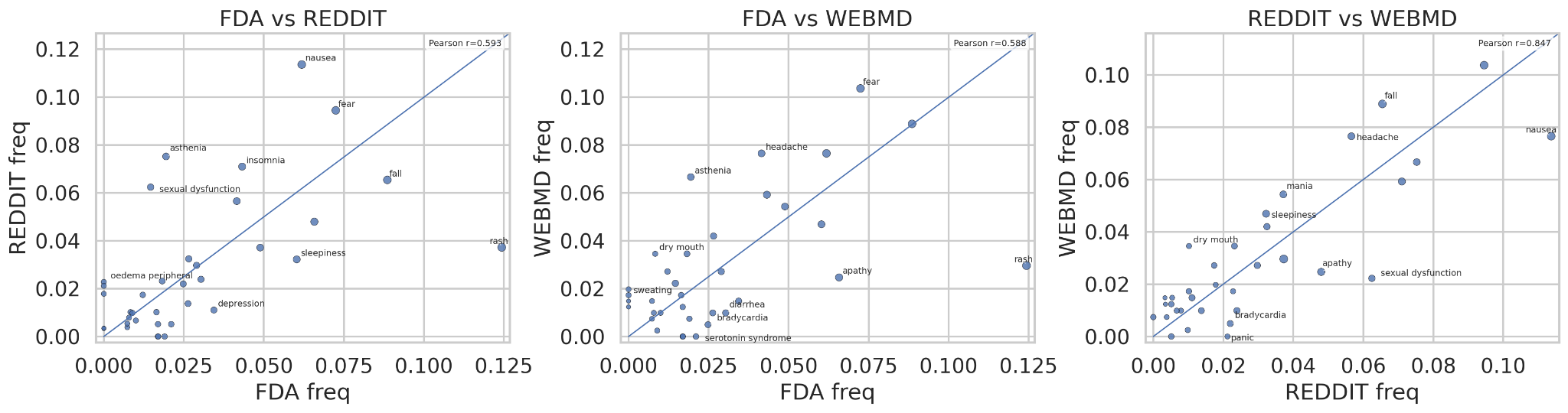}
  \caption{Pairwise scatterplots of normalized adverse-event frequencies for sertraline across
  FDA, Reddit, and WebMD. Each point represents one adverse event; the diagonal indicates equal
  relative frequency across the two compared sources.}
  \label{fig:ae-scatter}
\end{figure}

We next performed a focused analysis of sertraline. At the set level, sertraline followed the
same overall trend as the broader drug panel, with the greatest overlap between WebMD and Reddit
(Jaccard $= 0.786$), compared with FDA--WebMD ($0.577$) and FDA--Reddit ($0.558$).
Frequency-based comparisons were consistent with this pattern (Fig.~\ref{fig:ae-scatter}). In the sertraline scatterplots,
the correlation was moderate for FDA versus Reddit ($r = 0.593$) and FDA versus WebMD
($r = 0.588$), but substantially higher for Reddit versus WebMD ($r = 0.847$). The corresponding
dot-whisker plot showed the same ordering, indicating that the two community-derived sources were
more closely aligned not only in which AEs appeared, but also in their relative prominence.

Inspection of the sertraline scatterplots suggests that broadly recognizable, patient-salient
AEs---such as nausea, headache, sleepiness, rash, and sexual dysfunction---were shared across
sources and tended to appear among the more prominent signals. However, several points deviated
from the diagonal, indicating source-specific emphasis. In general, FDA-based comparisons appeared
to give relatively greater weight to more formally reported or medically coded events, whereas
Reddit and WebMD more strongly reflected symptoms that are directly felt in daily life and readily
discussed in lay narratives. This pattern is consistent with the stronger Reddit--WebMD
concordance observed in both overlap and frequency analyses.

\begin{figure}[htbp]
  \centering
  \includegraphics[width=0.95\linewidth]{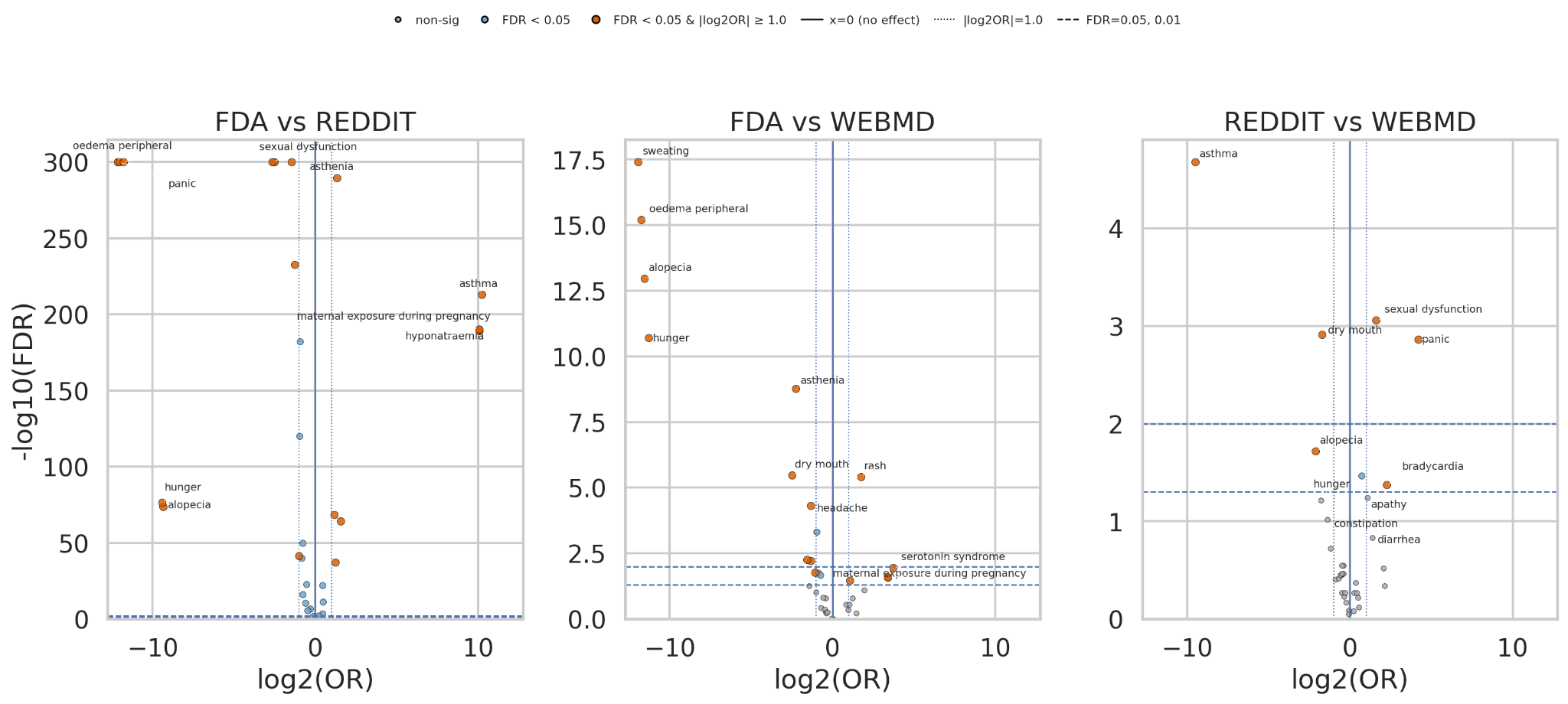}
  \caption{Volcano analysis of adverse-event frequency differences across data sources for sertraline. Each
  panel compares one source pair; the x-axis shows log$_2$ odds ratios and the y-axis shows
  $-\log_{10}$ false-discovery-rate-adjusted significance.}
  \label{fig:ae-volcano}
\end{figure}

To quantify source-specific enrichment in more detail, we examined pairwise volcano plots for
sertraline (Fig.~\ref{fig:ae-volcano}). These plots showed that most AEs clustered near the null region, consistent with a
shared core safety profile across sources, but a smaller subset showed clear source-preferential
enrichment. In the FDA-versus-Reddit comparison, several AEs were strongly shifted away from the
center, indicating substantial differences in relative representation between regulatory and
community reporting. In the FDA-versus-WebMD comparison, the degree of divergence was still
evident but generally less extreme. By contrast, the Reddit-versus-WebMD comparison showed fewer
large-effect outliers, again supporting the view that the two community-derived platforms are more
similar to one another than either is to FDA. Notably, some experiential symptoms such as dry
mouth, sexual dysfunction, and panic appeared more prominent in community-based comparisons,
whereas several medically framed events showed stronger enrichment in FDA, suggesting that source
differences reflect not only noise, but also distinct reporting incentives, symptom salience, and
coding practices.

\begin{figure}[htbp]
  \centering
  \includegraphics[width=0.95\linewidth]{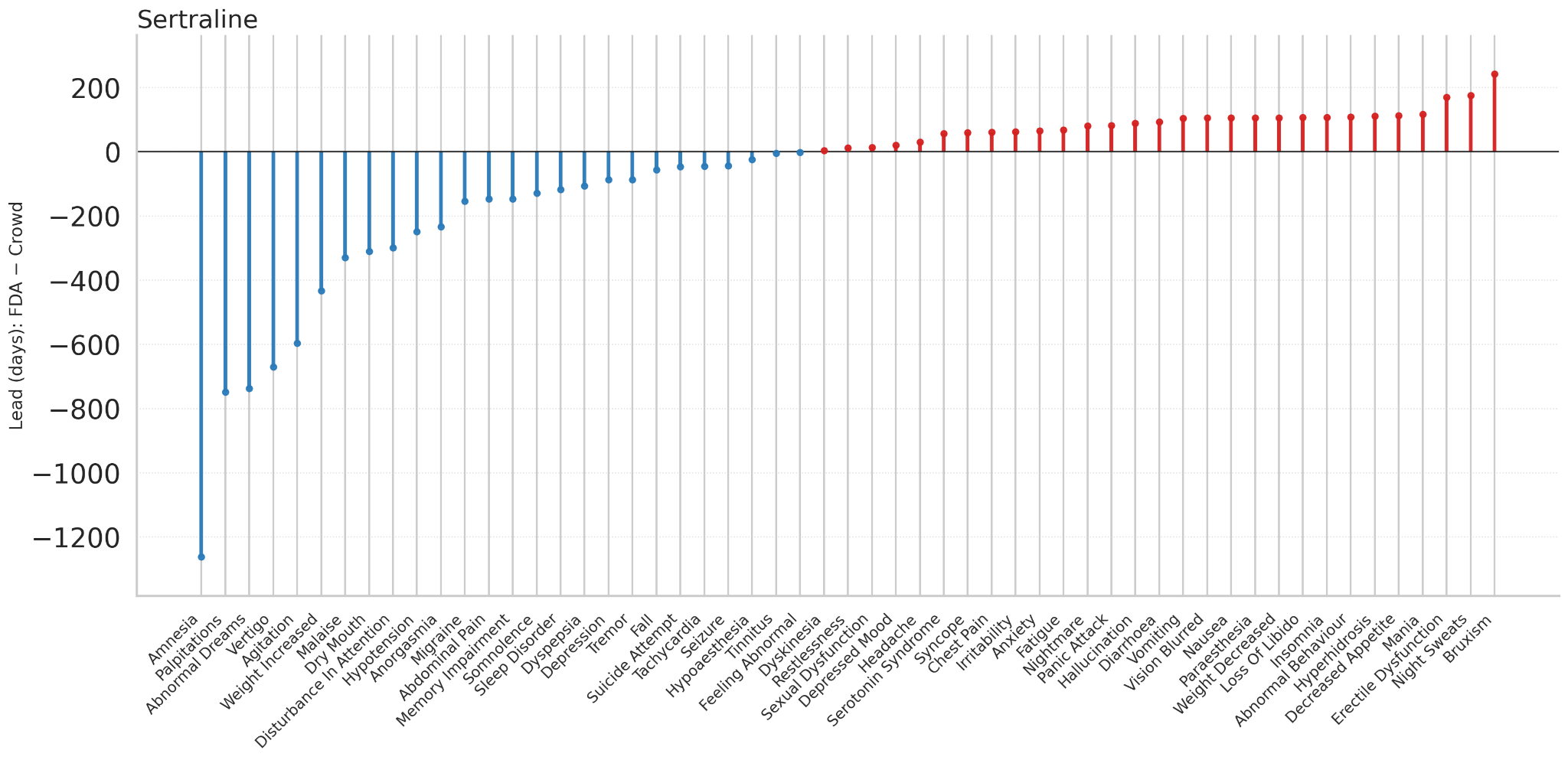}
  \caption{Longitudinal lead time of adverse-event first mention across surveillance streams for sertraline.
  Lead time (days) $=$ first FDA date $-$ min(first WebMD date, first Reddit date). Negative
  values indicate earlier mention in a community source; positive values indicate earlier
  appearance in FDA.}
  \label{fig:ae-leadtime}
\end{figure}

We further assessed temporal concordance using lead-time analysis for sertraline
(Fig.~\ref{fig:ae-leadtime}), defined as the
difference between the first FDA report date and the earliest first mention in the community sources.
Negative values therefore indicate that an AE was mentioned earlier in Reddit or WebMD than in FDA,
whereas positive values indicate earlier appearance in FDA. The lead-time distribution was skewed
toward negative values, with many sertraline AEs first appearing in community data substantially
earlier than in FDA, in some cases by several hundred days. This pattern suggests that community
platforms may capture a subset of patient-experienced AEs earlier than formal pharmacovigilance
channels. At the same time, a smaller set of AEs showed positive lead times, indicating earlier
appearance in FDA. These later-emerging community mentions may reflect events that become discussed only
after they are clinically recognized, formally coded, or amplified through broader public
awareness. Thus, the temporal analysis supports a complementary interpretation: community and
regulatory data do not simply duplicate one another, but may contribute different surveillance
timing for different types of AEs.

Taken together, these analyses indicate that FDA, WebMD, and Reddit capture a shared but
non-identical AE signal for antidepressants. Across drugs, the strongest similarity was usually
observed between the two community-derived platforms, whereas FDA remained related but
systematically distinct. The sertraline case study further showed that this distinction is evident
at multiple levels: AE set overlap, relative frequency structure, differential enrichment, and
time of first appearance. These findings support the use of a multi-source integration strategy,
in which regulatory data provide formal safety context and community sources add patient-centered
salience, narrative detail, and potentially earlier visibility for some adverse effects.

\begin{table}[htbp]
\centering
\caption{Pairwise adverse-event profile similarity and source balance across nine antidepressants.}
\label{tab:table4}
\small
\begin{tabular}{lccccc}
\toprule
\textbf{Drug} & \textbf{Jaccard} & \textbf{Jaccard} & \textbf{Jaccard} &
\textbf{Comp.} & \textbf{Evenness} \\
 & \textbf{(FDA--WebMD)} & \textbf{(FDA--Reddit)} & \textbf{(WebMD--Reddit)} &
 \textbf{Entropy} & \\
\midrule
Amitriptyline   & 0.509 & 0.471 & 0.723 & 0.424 & 0.386 \\
Desvenlafaxine  & 0.800 & 0.756 & 0.905 & 0.665 & 0.606 \\
Duloxetine      & 0.638 & 0.689 & 0.875 & 0.324 & 0.295 \\
Fluoxetine      & 0.580 & 0.529 & 0.833 & 0.726 & 0.661 \\
Paroxetine      & 0.612 & 0.540 & 0.805 & 0.476 & 0.433 \\
Phenelzine      & 0.610 & 0.549 & 0.523 & 0.758 & 0.690 \\
Sertraline      & 0.577 & 0.558 & 0.786 & 0.677 & 0.616 \\
Venlafaxine     & 0.592 & 0.667 & 0.660 & 0.518 & 0.471 \\
Vilazodone      & 0.286 & 0.756 & 0.293 & 0.468 & 0.426 \\
\bottomrule
\multicolumn{6}{p{0.92\linewidth}}{\footnotesize \textit{Note.} Pairwise Jaccard indices quantify
shared AE sets between sources. Composition Entropy (natural log scale) summarizes how evenly AE
counts are distributed across the three sources; Evenness ranges 0--1.}
\end{tabular}
\end{table}

\FloatBarrier

% ════════════════════════════════════════════════════════════════════════════
\section*{Discussion}
% ════════════════════════════════════════════════════════════════════════════

The multi-source pharmacovigilance and knowledge graph framework described here has three broad
implications for psychiatric medication information: it clarifies how community and regulatory
sources differ, it shows where community data may provide earlier contextual signals, and it
defines a retrieval architecture that keeps source provenance visible during response generation.

\paragraph{Community data as a complementary pharmacovigilance signal.}
FDA, WebMD, and Reddit did not provide interchangeable views of antidepressant AEs. The stronger
concordance between WebMD and Reddit (Jaccard similarity up to 0.905 for desvenlafaxine) than
between either community source and FAERS suggests that patient-generated data form a coherent,
partly independent signal. This does not make community data a substitute for regulatory
pharmacovigilance. Rather, it indicates that community platforms capture symptom salience,
everyday language, and reporting incentives that are less visible in formally coded safety
records. The source-balance results further show that this complementarity is compound-specific.
Some drugs showed relatively even cross-source representation, whereas others were concentrated
in one or two sources. Multi-source monitoring should therefore be interpreted at the
drug-specific level rather than treated as a uniform property of all antidepressants.

\paragraph{Temporal lead times and early signal context.}
The sertraline lead-time analysis suggests that community sources can contain AE mentions before
the corresponding FAERS receipt date, sometimes by several hundred days. This pattern supports
the use of social media and consumer reviews as supplementary signal contexts, but it should not
be read as proof that community platforms detect risk earlier in a causal or regulatory sense.
Lead time may reflect administrative delay, differences in coding practice, or the timing with
which patients choose to discuss symptoms publicly. The presence of positive lead times for some
AEs reinforces this boundary: community data are not uniformly earlier or more complete. Their
value lies in adding a patient-centred temporal layer to formal pharmacovigilance, especially
when interpreted alongside regulatory records rather than in isolation.

\paragraph{Knowledge graph architecture as provenance-aware retrieval.}
The knowledge graph addresses a practical problem for health-facing LLM systems: fluent answers
are not enough when the underlying evidence cannot be inspected. By grounding medications,
conditions, and adverse events in ATC-N, ICD-10, and MedDRA terms, and by preserving post-level
provenance through \textsc{MENTIONS} edges, the framework makes each retrieved claim traceable
to source-specific evidence. This design choice is also consistent with emerging evidence that clinical LLM evaluation should
move beyond isolated final-answer accuracy toward workflow-level, process-based assessment of
information gathering, reasoning, communication, and safety~\cite{ref20,ref21}. The multi-agent chatbot is therefore framed as an educational
retrieval and synthesis interface, not as an autonomous clinical decision system. This boundary
is especially important for psychiatric medication information, where unsupported statements
about discontinuation, suicidality, drug interactions, or vulnerable populations could increase
fear or distort treatment decisions. The architecture is designed to reduce hallucination risk by
separating evidence retrieval, cross-source comparison, and response validation, although its
actual safety and usefulness require prospective human evaluation.

\paragraph{Patient-facing interpretation.}
A central motivation for integrating community and regulatory data is that patients often need
both formal safety context and language that resembles lived experience. Reddit contributes broad
coverage of patient-described experiences, WebMD contributes semi-structured review context, and
FAERS contributes a formal post-marketing safety record. Keeping these roles separate within the
same retrieval framework may help future systems answer questions without flattening patient
narratives into regulatory facts or treating anecdotal reports as clinical incidence estimates.
This distinction is essential: community reports can make experiences easier to name and discuss,
but they cannot by themselves establish causality, prevalence, or individualized treatment risk.

\paragraph{Limitations.}
The NER pipeline achieved strong name-level F1 across entity types, but attribute-level
extraction was weaker, particularly for side-effect duration and frequency. Some adverse-event
characterizations in the graph may therefore be incomplete even when the entity name is correct.
The corpus is restricted to English-language posts and reviews, which may underrepresent
psychiatric medication experiences from non-English-speaking communities. Cross-source AE
harmonization relies on embedding-based matching to MedDRA and ICD-10. Although thresholds were
calibrated on physician-annotated data, residual mapping errors could affect similarity estimates.
The nine-antidepressant focus enables detailed multi-source comparison but limits immediate
generalizability to other drug classes. Finally, the current work evaluates data integration,
entity extraction, cross-source signal structure, and system architecture. It does not yet
establish clinical utility, response safety, patient usability, or effects on adherence or
decision-making. Future studies should extend the framework to broader medication classes, incorporate multilingual corpora, and evaluate chatbot responses prospectively with clinicians and patients before
deployment claims are made, ideally using workflow-level and safety-aware evaluation designs rather
than relying only on automated metrics or vignette-based testing~\cite{ref20,ref21}.

% ────────────────────────────────────────────────────────────────────────────
% Ethics Statement
% ────────────────────────────────────────────────────────────────────────────

% ────────────────────────────────────────────────────────────────────────────
% Data and Code Availability
% ────────────────────────────────────────────────────────────────────────────

% ────────────────────────────────────────────────────────────────────────────
% Acknowledgements
% ────────────────────────────────────────────────────────────────────────────

% ────────────────────────────────────────────────────────────────────────────
% Author Contributions
% ────────────────────────────────────────────────────────────────────────────

% ────────────────────────────────────────────────────────────────────────────
% Bibliography
% ────────────────────────────────────────────────────────────────────────────

\phantomsection
\label{sec:references}

% ════════════════════════════════════════════════════════════════════════════
\appendix
\renewcommand{\thesection}{\Alph{section}}
\setcounter{section}{0}
\setcounter{figure}{0}
\setcounter{table}{0}
\renewcommand{\thefigure}{S\arabic{figure}}
\renewcommand{\thetable}{S\arabic{table}}
% Appendix sections use bold with letter prefix, no rule
\titleformat{\section}{\large\bfseries}{Appendix~\thesection:}{0.5em}{}
% ════════════════════════════════════════════════════════════════════════════

\clearpage
\phantomsection
\section*{Appendices}
\vspace{1em}
\noindent{\large\bfseries Table of Contents\par}
\vspace{1.0em}
\appendixcontentsline{app:A}{A}{LLM Prompt for Expanding Generic Names to Brand Synonymy}
\appendixcontentsitem{tab:table5}{Table S1}{LLM prompt for expanding generic drug names to brand synonyms}
\appendixcontentsline{app:B}{B}{Binary Classification Model for Post Information Richness}
\appendixcontentsitem{tab:table6}{Table S2}{Binary classification performance for information richness filtering}
\appendixcontentsline{app:C}{C}{NER Annotation Interface}
\appendixcontentsitem{fig:annotation}{Figure S1}{NER annotation interface}
\appendixcontentsline{app:D}{D}{LLM Prompt for Named Entity Recognition}
\appendixcontentsline{app:E}{E}{Embedding Mapping Details and Cut-off Points}
\appendixcontentsitem{fig:thresholds-side-effects}{Figure S2}{ROC threshold calibration for side-effect mapping}
\appendixcontentsitem{fig:thresholds-conditions}{Figure S3}{ROC threshold calibration for condition mapping}
\appendixcontentsitem{fig:thresholds-comorbid}{Figure S4}{ROC threshold calibration for comorbid-condition mapping}
\appendixcontentsline{app:F}{F}{Cost Analysis for Model Comparison}
\appendixcontentsitem{tab:table8}{Table S3}{Efficiency and cost comparison of LLMs for the NER pipeline}
\appendixcontentsline{app:G}{G}{Knowledge Graph Construction Implementation Details}
\normalsize
\clearpage

\section{LLM Prompt for Expanding Generic Names to Brand Synonymy}
\label{app:A}

The brand-name expansion prompt was designed to return structured XML rather than free text
in order to simplify downstream parsing: a fixed tag schema
(\texttt{<generic>}/\texttt{<brand>}) ensures that each API response can be split without
regular-expression heuristics and is robust to variation in model phrasing. The phrase
``intended for nervous system related illnesses'' was included as a domain constraint to
prevent the model from returning homonymous brand names from other ATC classes (e.g.,
returning a cardiovascular trade name for a shared generic string). A
\texttt{<brand>None</brand>} sentinel was specified to distinguish true negatives (no marketed
brand found) from model refusals or off-topic outputs. The resulting XML was post-processed
by a tag-parser that extracted all \texttt{<brand>} values per generic, deduplicated across
API calls, and merged into the ATC-N dictionary as a curated synonym list used in all
downstream keyword-matching and entity-normalization steps.

\begin{table}[htbp]
\centering
\caption{LLM prompt for expanding generic drug names to brand synonyms.}
\label{tab:table5}
\small
\begin{tcolorbox}[colback=gray!5, colframe=gray!40, fontupper=\ttfamily\scriptsize,
  title=Brand-name expansion prompt, fonttitle=\bfseries\sffamily\small]
You are a drug brand name identifier. For each input generic drug name, return:\\[2pt]
\textless brand\_names\textgreater\\
\quad\textless generic\textgreater GENERIC\_NAME\textless /generic\textgreater\\
\quad\textless brand\textgreater BRAND\_NAME\_1\textless /brand\textgreater\\
\quad\textless brand\textgreater BRAND\_NAME\_2\textless /brand\textgreater\\
\textless /brand\_names\textgreater\\[4pt]
If no brand names found, return:\\[2pt]
\textless brand\_names\textgreater\\
\quad\textless generic\textgreater GENERIC\_NAME\textless /generic\textgreater\\
\quad\textless brand\textgreater None\textless /brand\textgreater\\
\textless /brand\_names\textgreater\\[4pt]
Only return the XML. No explanations. The drugs should be intended for nervous system related
illnesses.\\[4pt]
Input drug generic name: \{generic\_name\}
\end{tcolorbox}
\end{table}

\FloatBarrier
\clearpage

\section{Binary Classification Model for Post Information Richness}
\label{app:B}

During exploratory inspection, we observed that many Reddit posts still lacked substantive content
after initial keyword-based and rule-based filtering. To further improve data quality, we
implemented a secondary data cleaning stage aimed at distinguishing \textit{information-rich} from
\textit{information-poor} posts. We first randomly sampled 3,500 posts from the cleaned Reddit
corpus and manually labeled them according to three criteria:

\begin{enumerate}
  \item The post is about a Nervous System drug (e.g., antidepressants, antipsychotics, mood
  stabilizers, anxiolytics).
  \item The post mentions a side effect or adverse event related to the drug. A \emph{side effect}
  is a common or expected reaction (e.g., fatigue, weight gain); an \emph{adverse event} is a
  severe or unexpected reaction (e.g., seizures, suicidal thoughts).
  \item The post is \emph{information-rich}, meaning it includes at least one of the following:
  specific symptom(s) or effect(s); details about timing, duration, dosage, or sequence of events;
  or how the reaction impacted daily life or required medical attention.
\end{enumerate}

Approximately 23.3\% of posts were labeled as information-rich. Using these labeled data, we then
fine-tuned a \textit{BERT-base-uncased} model for binary classification using the Hugging Face
Transformers (v4.52.3) library and PyTorch. The model architecture comprised 12 transformer
layers, each with 12 self-attention heads and a hidden size of 768, totaling approximately
110~million parameters. Key hyperparameters included a hidden dropout probability of 0.1,
intermediate layer size of 3,072, GELU activation, and maximum sequence length of 512 tokens.
Training used a batch size of 16, learning rate of $2 \times 10^{-5}$, and three epochs,
optimized with AdamW and float32 precision.

After three epochs, the classifier achieved weighted accuracy of 0.866, with precision 0.875,
recall 0.866, and F1 score 0.869 on the test set. The trained model was then used to
automatically filter the remaining Reddit corpus, retaining only posts predicted as
information-rich for subsequent analyses. Using the pre-training classifier, we filtered 466,525
information-rich posts from the original 1,138,331 posts. To assess post-filtering quality, we
randomly sampled 1,000 retained posts and manually verified that the majority contained
substantive narratives regarding psychiatric medication use and side effects.

\begin{table}[htbp]
\centering
\caption{Binary classification performance for information richness filtering.}
\label{tab:table6}
\small
\begin{tabular}{lcccc}
\toprule
\textbf{Class} & \textbf{Support} & \textbf{Precision} & \textbf{Recall} & \textbf{F1} \\
\midrule
Not Information Rich & 349 & 0.93 & 0.87 & 0.90 \\
Information Rich     & 151 & 0.74 & 0.85 & 0.79 \\
\bottomrule
\end{tabular}
\end{table}

\clearpage

\section{NER Annotation Interface}
\label{app:C}

\begin{figure}[H]
  \centering
  \includegraphics[width=0.95\linewidth]{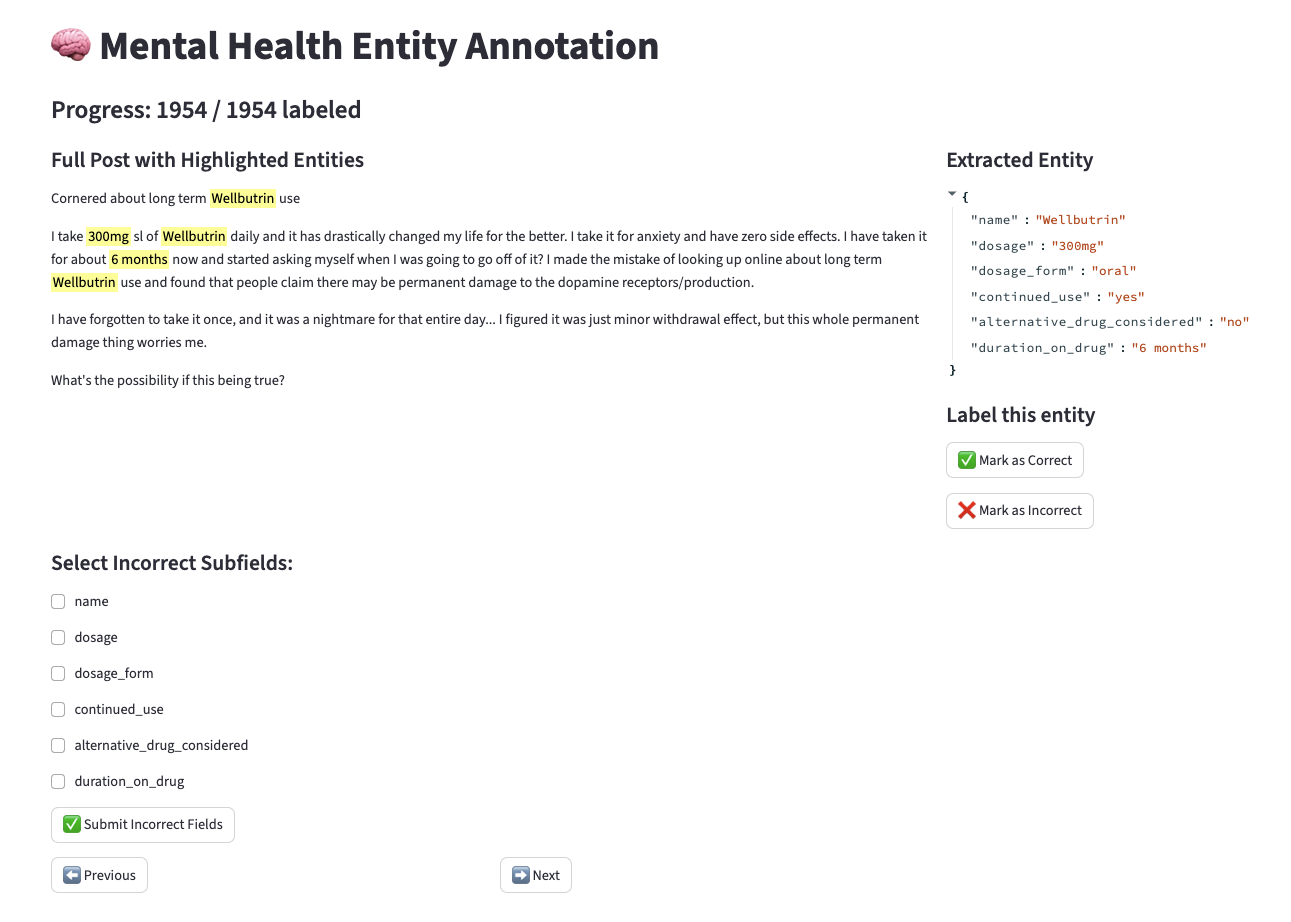}
  \caption{Mental-health entity annotation interface for medical doctors. The interface shows a
  full Reddit post on the left with model-highlighted spans. The right panel presents the
  structured extraction as a JSON preview. Medical doctors either confirm the extraction or flag
  errors, and can pinpoint specific subfields via checkboxes before submitting corrections.}
  \label{fig:annotation}
\end{figure}

\FloatBarrier

\clearpage

\section{LLM Prompt for Named Entity Recognition}
\label{app:D}

The NER prompt was designed as a single-pass structured extraction: all entity types
(drugs, condition, comorbid conditions, side effects) and their pairwise relations are
extracted in one model call per post, reducing API cost compared with sequential
entity-specific prompts. A JSON output schema was selected over XML because it maps
directly to Python dictionaries and natively supports the nested objects and arrays required
for multi-drug and multi-side-effect posts. A strict null-handling rule
(``all missing fields \textsc{must} be null'') was specified to prevent the model from
hallucinating absent attributes; this was especially important for dosage, duration, and
severity fields, which patients rarely state explicitly. Sentiment
(\texttt{positive}/\texttt{negative}/\texttt{neutral}) and effectiveness were added as
auxiliary labels to enable post-level stratification in downstream analyses without
requiring a separate classification call. The instruction ``Respond with ONLY the JSON
output'' and the prohibition on paraphrasing were included to suppress the conversational
preamble that LLMs otherwise produce, which would break automated JSON parsing.

\begin{tcolorbox}[breakable, colback=gray!5, colframe=gray!40, fontupper=\ttfamily\scriptsize,
  title=Prompt used for named entity recognition, fonttitle=\bfseries\sffamily\small]
prompt = f"""\\[4pt]
You are a clinical language evaluator and medical NLP assistant. Your task is to extract
structured mental health-related information from user-generated Reddit content to support
knowledge graph construction in psychiatry and psychology.\\[4pt]
From the given post, extract the following information in JSON format.\\[4pt]
\textbf{Required entities and attributes:}\\
- drugs: a list of drug objects, each with: name, dosage, dosage\_form,
continued\_use (yes/no), alternative\_drug\_considered (yes/no),
duration\_on\_drug\\
- condition: name, severity (if explicitly mentioned), duration (if explicitly mentioned),
diagnosed (yes/no/unknown)\\
- comorbid\_conditions: list of other explicitly mentioned psychiatric or psychological
conditions (e.g., ADHD, anxiety, PTSD)\\
- side\_effects: name, severity, frequency, duration, associated\_drug\\[4pt]
\textbf{Strict guidelines:}\\
1. \textbf{Medication Identification}: Extract psychiatric or psychological medications only if
explicitly mentioned (e.g., SSRIs like Prozac, antipsychotics like Seroquel, mood stabilizers
like Lamictal). Infer dosage\_form as ``oral'' unless explicitly stated otherwise (e.g.,
injection). Set continued\_use to ``yes'' if the user mentions they are still taking it, and
``no'' if they have stopped or switched. Set alternative\_drug\_considered to ``yes'' if
another drug is discussed for switching or comparing; otherwise set it to ``no''. Extract
duration\_on\_drug only when the time taking the drug is clearly stated (e.g., ``for 3 days'',
``took it for 2 months''); otherwise set it to null.\\[3pt]
2. \textbf{Condition Identification}: Only extract explicitly named psychiatric or psychological
conditions (e.g., depression, bipolar disorder, PTSD). Extract severity (e.g., ``mild'',
``severe'', ``moderate'') only if mentioned. Set diagnosed to ``yes'' if the post mentions a
formal diagnosis, ``no'' if the user says self-diagnosed, and ``unknown'' if unclear. Extract
duration only if a time period is clearly described (e.g., ``struggled for 3 years'').\\[3pt]
3. \textbf{Comorbid Conditions}: Extract all other named mental or behavioral health conditions
in the post beyond the primary one. Only extract conditions explicitly stated (e.g., ``I have
ADHD and anxiety'').\\[3pt]
4. \textbf{Side Effects}: Only extract when clearly described and attributed to a drug. All fields
(severity, frequency, duration) must be explicitly mentioned; otherwise return null. Associate
side effects to drugs if stated (e.g., ``Lexapro gave me nausea''). If a side effect is clearly
caused by stopping or tapering a drug, the relationship is ``causes\_by\_withdraw''; otherwise
the relationship is ``causes''.\\[3pt]
5. \textbf{Null Handling}: All missing or unmentioned fields MUST be null. Use empty arrays if no
side effects, therapies, or comorbid conditions are mentioned. Omit entire objects (e.g., drug,
side effects) if nothing is mentioned.\\[3pt]
6. \textbf{Output Format}: Strictly adhere to the JSON format below. No explanations or inferred
data.\\[3pt]
7. \textbf{Sentiment and Effectiveness}: Overall sentiment should be ``positive'', ``negative'',
or ``neutral''. Only categorize effectiveness or sentiment if explicitly stated.\\[4pt]
\textbf{JSON output format:}\\
\{\\
\phantom{xx}"structured\_info": \{\\
\phantom{xxxx}"drugs": [\{\\
\phantom{xxxxxx}"name": "",\\
\phantom{xxxxxx}"dosage": "",\\
\phantom{xxxxxx}"dosage\_form": "",\\
\phantom{xxxxxx}"continued\_use": "",\\
\phantom{xxxxxx}"alternative\_drug\_considered": "",\\
\phantom{xxxxxx}"duration\_on\_drug": ""\\
\phantom{xxxx}\}],\\
\phantom{xxxx}"condition": \{\\
\phantom{xxxxxx}"name": "",\\
\phantom{xxxxxx}"severity": "",\\
\phantom{xxxxxx}"duration": "",\\
\phantom{xxxxxx}"diagnosed": ""\\
\phantom{xxxx}\},\\
\phantom{xxxx}"comorbid\_conditions": [],\\
\phantom{xxxx}"side\_effects": [\{\\
\phantom{xxxxxx}"name": "",\\
\phantom{xxxxxx}"severity": "",\\
\phantom{xxxxxx}"frequency": "",\\
\phantom{xxxxxx}"duration": "",\\
\phantom{xxxxxx}"associated\_drug": ""\\
\phantom{xxxx}\}]\\
\phantom{xx}\},\\
\phantom{xx}"relations": [\\
\phantom{xxxx}\{ "start": \{ "label": "Medication", "properties": \{ "name": "" \} \},
"end": \{ "label": "Condition", "properties": \{ "name": "" \} \},
"relation": "treats", "properties": \{ "diagnosed": null, "off\_label": null \} \},\\
\phantom{xxxx}\{ "start": \{ "label": "Medication", "properties": \{ "name": "" \} \},
"end": \{ "label": "SideEffect", "properties": \{ "name": "" \} \},
"relation": "causes", "properties": \{ "severity": null, "duration": null, "dosage": null \} \},\\
\phantom{xxxx}\{ "start": \{ "label": "Medication", "properties": \{ "name": "" \} \},
"end": \{ "label": "SideEffect", "properties": \{ "name": "" \} \},
"relation": "causes\_by\_withdraw", "properties": \{\} \},\\
\phantom{xxxx}\{ "start": \{ "label": "Condition", "properties": \{ "name": "" \} \},
"end": \{ "label": "Condition", "properties": \{ "name": "" \} \},
"relation": "comorbid\_with", "properties": \{\} \}\\
\phantom{xx}],\\
\phantom{xx}"sentiment": "",\\
\phantom{xx}"effectiveness": ""\\
\}\\[4pt]
Text to Analyze: \{text\}\\[4pt]
Respond with ONLY the JSON output. Do not paraphrase or reword any content; extract the
information exactly as written in the original post.\\[4pt]
"""
\end{tcolorbox}

\FloatBarrier

\clearpage

\section{Embedding Mapping Details and Cut-off Points}
\label{app:E}

To translate free-text entities into controlled vocabularies, we used embedding-based
nearest-neighbour matching and selected operating thresholds empirically against a
physician-annotated gold standard. Specifically, we first generated a gold NER set by dual human
adjudication and consensus, then embedded each extracted entity string (lower-cased and trimmed)
with \textit{text-embedding-3-small}. For each entity, we computed cosine similarity to all
reference terms and retained the single most similar candidate (top-1). We labeled a match as
``correct'' when either (i)~the top-1 canonical term was exactly equal to the normalized gold
string or (ii)~manual adjudication marked the pair as semantically the same.
These binary labels provided ground truth for threshold selection.

Using these labels, we evaluated similarity thresholds by sweeping the decision boundary over the
continuous similarity score and computing precision--recall (PR) curves, receiver operating
characteristic (ROC) curves, and the area under the ROC curve (AUC). We then identified the
operating point that maximized Youden's J statistic (TPR $-$ FPR), which balances sensitivity
against specificity and avoids overfitting to prevalence. We confirmed that the Youden-selected
cutoffs aligned with the ``elbow'' region of the PR curves, providing a practical trade-off
between false merges (over-mapping) and misses (under-mapping).

Because the distribution of similarity scores differed by entity type, we tuned thresholds
separately for side effects (mapped to MedDRA Preferred Terms) and for conditions and comorbid
conditions (mapped to ICD-10 terms). The final operating thresholds used throughout the study
were:
\begin{itemize}
  \item Side effects $\rightarrow$ MedDRA PT: $\tau = 0.68$
  \item Condition $\rightarrow$ ICD-10: $\tau = 0.56$
  \item Comorbid condition $\rightarrow$ ICD-10: $\tau = 0.54$
\end{itemize}

\begin{figure}[htbp]
  \centering
  \includegraphics[width=0.95\linewidth]{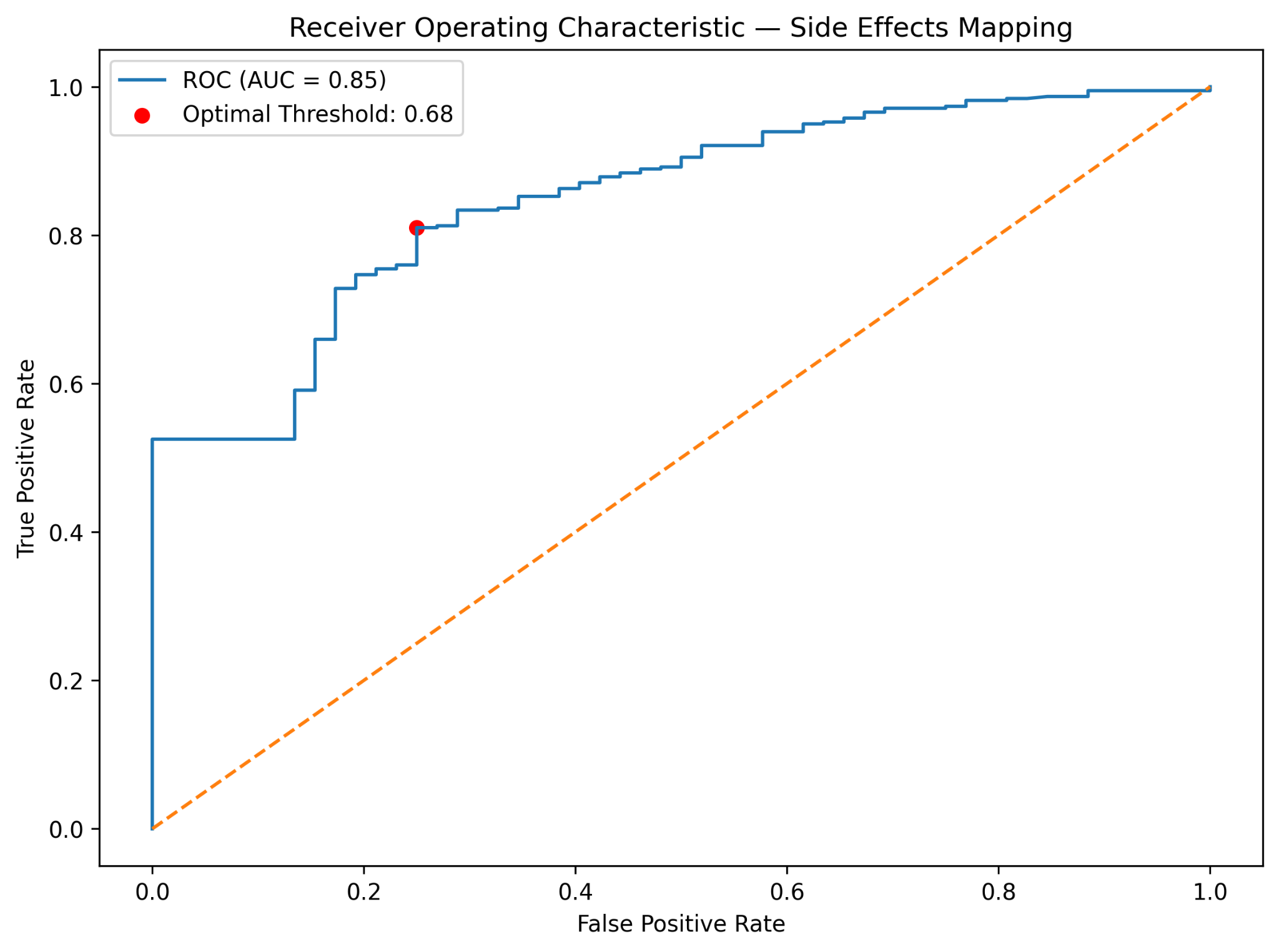}
  \caption{ROC threshold calibration for side-effect mapping. The red point indicates the selected
  Youden operating point ($\tau = 0.68$).}
  \label{fig:thresholds-side-effects}
\end{figure}

\begin{figure}[htbp]
  \centering
  \includegraphics[width=0.95\linewidth]{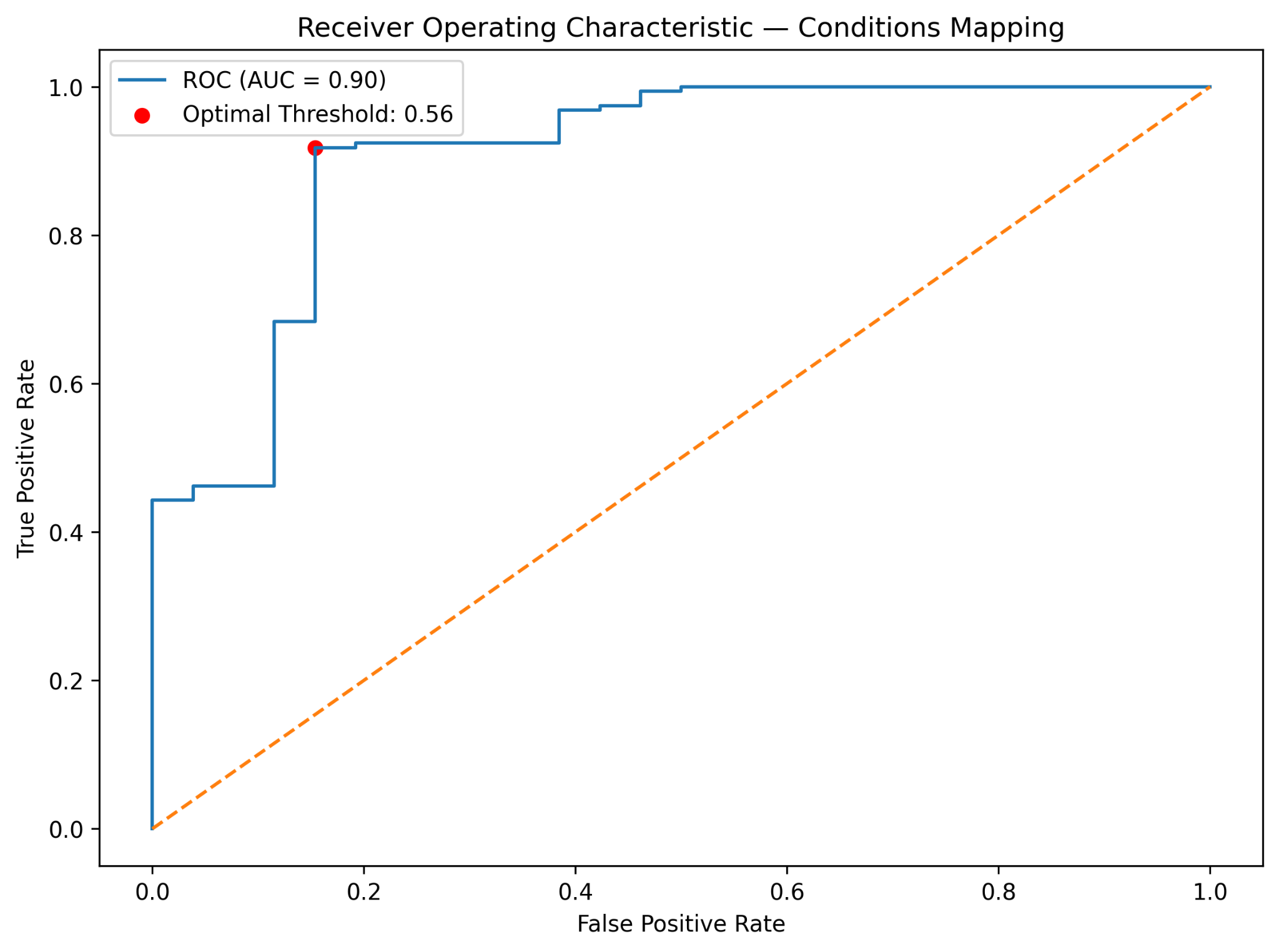}
  \caption{ROC threshold calibration for condition mapping. The red point indicates the selected
  Youden operating point ($\tau = 0.56$).}
  \label{fig:thresholds-conditions}
\end{figure}

\begin{figure}[htbp]
  \centering
  \includegraphics[width=0.95\linewidth]{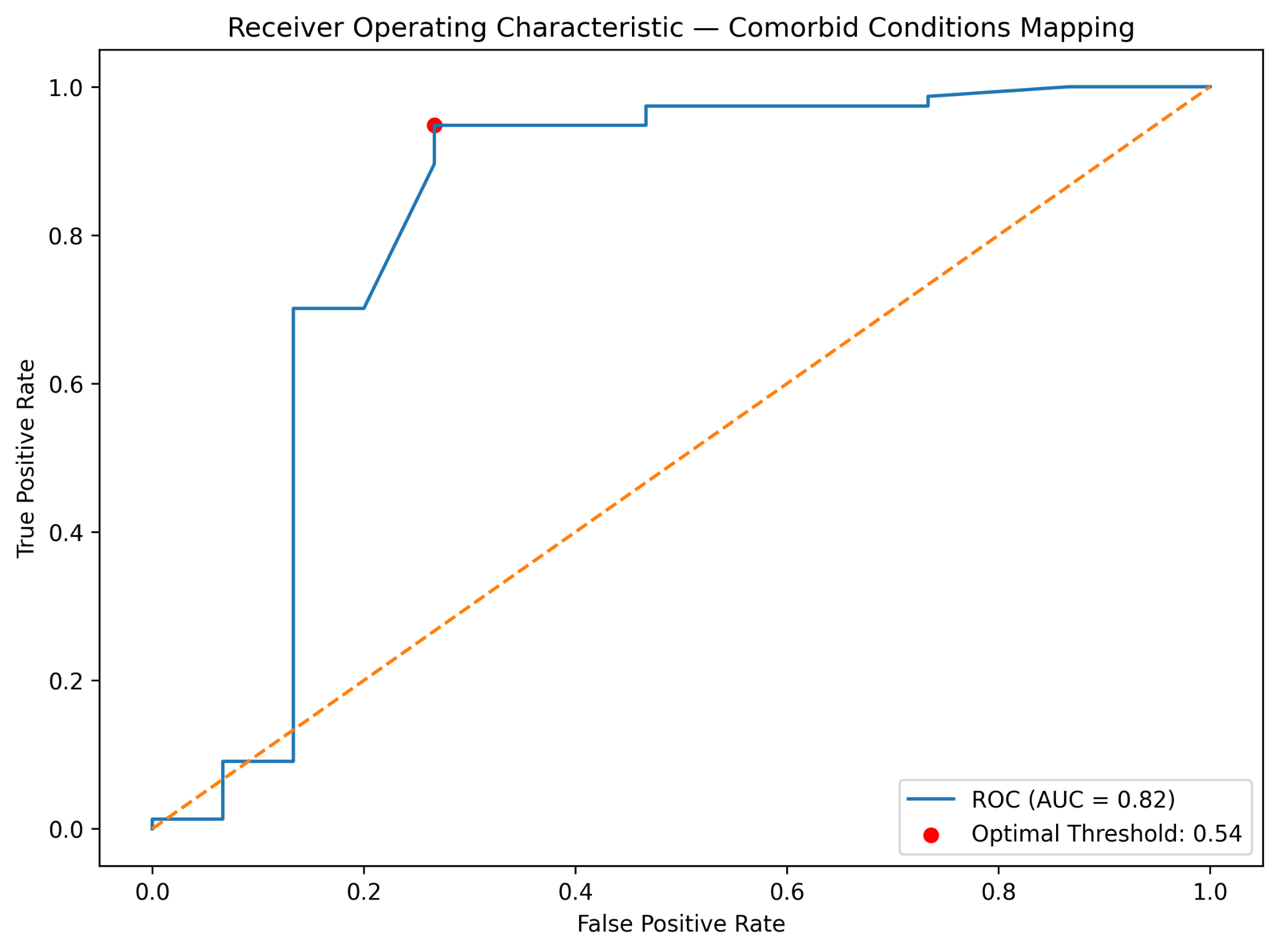}
  \caption{ROC threshold calibration for comorbid-condition mapping. The red point indicates the
  selected Youden operating point ($\tau = 0.54$).}
  \label{fig:thresholds-comorbid}
\end{figure}

These cutoffs were chosen on the labeled validation subset (derived from the physician gold NER)
and then applied to the full corpus. For transparency and reproducibility,
Figs.~\ref{fig:thresholds-side-effects}--\ref{fig:thresholds-comorbid} show ROC plots generated
from the labeled data to illustrate performance across the full range of similarity values.

\FloatBarrier
\clearpage

\section{Cost Analysis for Model Comparison}
\label{app:F}

In this section, we compare multiple LLMs under a fixed NER prompt/schema to identify a
configuration that scales to our full corpus with acceptable accuracy and cost. We quantify
efficiency (tokens, latency, wall-clock time) and operational spend (per-question and total), and
interpret these alongside the main-text NER quality metrics. The goal is a practical default model
that minimizes downstream adjudication while remaining fast and budget-conscious for repeated
runs.

With average inputs essentially constant ($\approx$1.5k tokens), differences arise from output
length, latency, and pricing. GPT-4.1-nano is the fastest ($\approx$0.041~s/request;
$\approx$5.4~h total) and among the cheapest (\$0.0029/question; \$188 total), but its lower
capacity corresponds to weaker extractions on harder posts in our evaluation. GPT-5-nano is the
least expensive overall (\$0.0024/question; \$156 total) but slower in aggregate ($\approx$22.5~h).
Premium models, notably Claude-Sonnet-4, are competitive on latency ($\approx$0.069~s;
$\approx$9~h total) but prohibitively costly for repeated runs (\$0.0768/question; \$4,991 total).
Mid-tier options such as GPT-4o-mini (\$0.0042/question; \$270; $\approx$7.3~h) and
Deepseek-V3 (\$0.0054/question; \$349; $\approx$18.1~h) offer moderate spend and speed.
Gemini-2.5-Flash and GPT-5-mini land near GPT-4.1-mini on per-question price but are slower or
costlier in total for our workload.

Balancing accuracy, speed, and cost, GPT-4.1-mini is the preferred default for this study. It
maintains low latency ($\approx$0.066~s/request; $\approx$8.6~h total) and a moderate budget
(\$0.012/question; \$777 total), while delivering stronger entity and attribute extraction than
the nano tier on long, multi-entity posts. In practice, this balance reduces manual correction
without incurring the steep costs of premium models, making GPT-4.1-mini the most efficient and
reproducible choice for our NER pipeline.

\begin{table}[htbp]
\centering
\caption{Efficiency and cost comparison of LLMs for the NER pipeline.}
\label{tab:table8}
\small
\begin{tabular}{lrrrrr}
\toprule
\textbf{Model} & \textbf{Avg.\ Output} & \textbf{Avg.\ Time/} & \textbf{Cost/} &
\textbf{Est.\ Total} & \textbf{Est.\ Total} \\
 & \textbf{Tokens} & \textbf{Request (s)} & \textbf{Question (\$)} &
\textbf{Cost (\$)} & \textbf{Time (h)} \\
\midrule
GPT-5-mini        & 701.7 & 0.184 & 0.0125 &   811 & 23.9 \\
GPT-5-nano        & 477.0 & 0.174 & 0.0024 &   156 & 22.5 \\
\textbf{GPT-4.1-mini} & \textbf{687.3} & \textbf{0.066} & \textbf{0.0120} & \textbf{777} & \textbf{8.6} \\
GPT-4.1-nano      & 651.2 & 0.041 & 0.0029 &   188 &  5.4 \\
GPT-4o-mini       & 614.7 & 0.057 & 0.0042 &   270 &  7.3 \\
Claude-Sonnet-4   & 720.2 & 0.069 & 0.0768 & 4,991 &  9.0 \\
Gemini-2.5-Flash  & 815.6 & 0.158 & 0.0123 &   801 & 20.4 \\
Deepseek-V3       & 740.6 & 0.140 & 0.0054 &   349 & 18.1 \\
Qwen3-235b-A22b   & 880.2 & 0.211 & 0.0071 &   459 & 27.3 \\
\bottomrule
\multicolumn{6}{p{0.9\linewidth}}{\footnotesize \textbf{Bold} row = selected pipeline default
(GPT-4.1-mini). Costs are estimated from published API pricing at the time of the study.}
\end{tabular}
\end{table}

\FloatBarrier

\clearpage

\section{Knowledge Graph Construction Implementation Details}
\label{app:G}

We constructed a Reddit-based mental health knowledge graph (KG) by ingesting a JSON corpus of
posts previously processed by an NLP pipeline to extract entities and relations. Each JSON record
corresponded to a single Reddit post and contained: (i)~a unique \texttt{post\_id} and optional
index; (ii)~the original post text; (iii)~auxiliary model outputs (e.g., sentiment and
effectiveness scores); and (iv)~a \texttt{structured\_info} object listing extracted medications,
conditions, comorbid conditions, side effects, and pairwise relations between them.

\subsection*{Schema and Canonicalization}

The KG was implemented in Neo4j with four main node types: Post, Medication, Condition, and
SideEffect. We treated Post nodes as lightweight anchor
nodes containing only a unique identifier (\texttt{post\_id}) and minimal metadata (e.g., an index
and creation timestamp); post text and model outputs were not stored in the graph to keep the KG
compact and privacy-preserving.

Domain entities were canonicalized before insertion. For medications, we required the presence of
a canonical ingredient-level name (\texttt{level\_5\_name}) produced by an upstream mapping step;
records without this field were discarded. The canonical name was stored as the node's
\texttt{name}, and temporary mapping fields (e.g., similarity scores) were removed. Conditions were
required to have an \texttt{icd10-term} field and a non-missing similarity score indicating a
successful mapping; the node \texttt{name} was set to this ICD-10 term and the similarity value was
dropped from the stored properties. Side effects were processed analogously, retaining only entries
with a non-null similarity score and a valid \texttt{meddra-term}, which became the node
\texttt{name}. For each entity, we derived a deterministic, lowercased, whitespace-normalized unique
identifier (\texttt{uid}) of the form \texttt{med::...}, \texttt{cond::...}, or \texttt{se::...}.
Neo4j uniqueness constraints on \texttt{uid} (for \texttt{Medication}, \texttt{Condition}, and
\texttt{SideEffect}) and on \texttt{post\_id} (for \texttt{Post}) ensured that logically identical
entities were merged across posts.

\subsection*{Relations and Evidence Aggregation}

We modeled three classes of edges. First, we created post-to-entity mention edges
(\texttt{Post})-[\textsc{:MENTIONS}]->(\texttt{Medication}|\texttt{Condition}|\texttt{SideEffect})
for every canonical entity detected in a post. These edges provide a direct trace from graph
entities back to the user-generated content in which they appear.

Second, we imported typed entity--entity relations derived from the NLP pipeline. Each relation in
the JSON input specified a start node, end node, and a relation label (e.g., \texttt{treats},
\texttt{causes}, \texttt{causes\_by\_withdraw}, \texttt{comorbid\_with}). We mapped these labels to
a fixed set of Neo4j relationship types: \textsc{TREATS}, \textsc{CAUSES},
\textsc{CAUSES\_BY\_WITHDRAW}, and \textsc{COMORBID\_WITH}. For each endpoint, we reused the same
canonicalization logic as for standalone entities; if a canonicalized entity did not yet exist in
the batch, it was created and linked back to the corresponding posts via \textsc{MENTIONS}.
Relation-level properties (such as severity, duration, dosage, and mapped name fields) were
retained, whereas similarity scores used only for upstream mapping were stripped.

To support evidence aggregation, each unique triple (\texttt{from\_uid}, \texttt{relation\_type},
\texttt{to\_uid}) was treated as a single relationship instance in the graph and attached to a list
of supporting \texttt{post\_ids}. During ingestion, if a relation between two entities already
existed, the current \texttt{post\_id} was appended to this list only if it was not already present,
thereby accumulating the set of posts that mention the same entity pair and relation type.

Third, we automatically constructed comorbidity edges between the primary condition and each
comorbid condition mentioned in a post. Specifically, when a post contained a primary condition plus
one or more comorbid conditions in the structured information, we added \textsc{COMORBID\_WITH}
edges from the primary condition to each comorbid condition, regardless of whether an explicit
comorbidity relation had been generated by the NLP model. This ensured consistent representation of
comorbid patterns even when relation extraction was conservative.

\subsection*{Batched Ingestion and Sidecar Database}

To scale to the full Reddit corpus, we implemented a batched ingestor in Python using the Neo4j
Bolt driver. The script streams the JSON file either as an array or as JSONL, groups records into
batches (typically $10^{3}$ posts), and, for each batch, constructs in-memory collections of posts,
canonicalized entities, mention edges, and relations. These are written to Neo4j in a single
transactional \texttt{UNWIND} call per node/edge type, using \texttt{MERGE} semantics so that
entities and relationships are deduplicated across batches while their evidence (\texttt{post\_ids})
is incrementally updated.

Because we intentionally keep the graph compact, we maintain a separate SQLite ``sidecar''
database that stores the full post text, sentiment and effectiveness outputs, and per-post evidence
tables linking posts to entity UIDs and relation instances. The sidecar uses a full-text search
index over post text and indexed foreign keys to entity and relation tables, enabling efficient
retrieval of raw textual context and model annotations for any subset of graph nodes or edges
without polluting the core KG with unstructured text.

\end{document}